\documentclass[journal]{IEEEtran}
\usepackage{graphicx}
\usepackage{multirow}
\usepackage{epstopdf}
\usepackage{times}
\usepackage{amsmath}
\usepackage{amssymb}
\usepackage{amsfonts}
\usepackage{color}
\usepackage{subfigure}
\usepackage[colorlinks]{hyperref}
 \usepackage{colortbl}
\usepackage[ruled, linesnumbered, vlined]{algorithm2e}
\usepackage{algorithmic}
\usepackage{rotating}
\usepackage{adjustbox,lipsum}
\usepackage{dsfont}
\usepackage{booktabs}
\usepackage{multirow}
\usepackage{array}

\usepackage{pifont}

\newcommand{\etal}{{\em et al.\,}}       
\newcommand{\eg}{{\em e.g.}}           
\newcommand{\ie}{{\em i.e.}}           

\newcommand{\bl}[1]{\textcolor{blue}{#1}}

\begin{document}

\title{NormAUG: Normalization-guided Augmentation for Domain Generalization}

\author{Lei Qi,
        Hongpeng Yang,
        Yinghuan Shi,
        Xin Geng
\thanks{The work is supported by NSFC Program (Grants No. 62206052, 62125602, 62076063), Jiangsu Natural Science Foundation Project (Grant No. BK20210224), and the Xplorer Prize.}
\thanks{Lei Qi is with the School of Computer Science and Engineering, Southeast University, National Center of Technology Innovation for EDA, and Key Laboratory of New Generation Artificial Intelligence Technology and Its Interdisciplinary Applications (Southeast University), Ministry of Education, China, 211189 (e-mail: qilei@seu.edu.cn).}
\thanks{Hongpeng Yang is with the School of Cyber Science and Engineering,  Southeast University, Nanjing, China, 211189 (e-mail: hp\_yang@seu.edu.cn).}
\thanks{Yinghuan Shi is with the State Key Laboratory for Novel Software Technology, Nanjing University, Nanjing, China, 210023 (e-mail: syh@nju.edu.cn).}
\thanks{Xin Geng is with the School of Computer Science and Engineering, Southeast University, and Key Laboratory of New Generation Artificial Intelligence Technology and Its Interdisciplinary Applications (Southeast University), Ministry of Education, China, 211189 (e-mail: xgeng@seu.edu.cn).}
\thanks{Corresponding author: Xin Geng.}
}

%
%

\markboth{ }%
{Shell \MakeLowercase{\textit{et al.}}: Bare Demo of IEEEtran.cls for IEEE Journals}

\maketitle

\begin{abstract}
Deep learning has made significant advancements in supervised learning. However, models trained in this setting often face challenges due to domain shift between training and test sets, resulting in a significant drop in performance during testing. To address this issue, several domain generalization methods have been developed to learn robust and domain-invariant features from multiple training domains that can generalize well to unseen test domains. Data augmentation plays a crucial role in achieving this goal by enhancing the diversity of the training data.
In this paper, inspired by the observation that normalizing an image with different statistics generated by different batches with various domains can perturb its feature, we propose a simple yet effective method called NormAUG (Normalization-guided Augmentation). Our method includes two paths: the main path and the auxiliary (augmented) path. During training, the auxiliary path includes multiple sub-paths, each corresponding to batch normalization for a single domain or a random combination of multiple domains. This introduces diverse information at the feature level and improves the generalization of the main path.
Moreover, our NormAUG method effectively reduces the existing upper boundary for generalization based on theoretical perspectives. During the test stage, we leverage an ensemble strategy to combine the predictions from the auxiliary path of our model, further boosting performance. Extensive experiments are conducted on multiple benchmark datasets to validate the effectiveness of our proposed method.
\end{abstract}


%
\IEEEpeerreviewmaketitle

\section{Introduction}


\IEEEPARstart{I}{n} the last decade, deep learning has achieved significant success in various applications, including classification~\cite{DBLP:conf/cvpr/HeZRS16,aversa2020deep}, object detection~\cite{DBLP:conf/iccv/0066HLY0GD21}, and semantic segmentation~\cite{DBLP:conf/miccai/RonnebergerFB15,lingwal2023semantic}. Most existing methods are based on the assumption of iid (independent and identically distributed) data, where the training and test data are assumed to be from the same distribution. However, in real-world applications, this assumption is often violated due to domain shift between the training (source) and test (target) domains. For instance, a model trained on photo images may not perform well on sketch images due to distributional variations, \ie, domain generalization (DG). In the DG task, the unknown data distribution of the test data poses a challenge. If the training samples lack diversity, the trained model may overfit to the training data, hindering its generalization ability. Enriching the diversity of training data can be viewed as a way to simulate the distribution of diverse data, allowing the model to capture the characteristics of unseen test data during the training stage. It is worth noting that perturbing features is a technique employed to enhance the diversity of training data, as demonstrated in various studies~ \cite{DBLP:conf/iccv/LiLLGFH21,Li_2023_ICCV,wang2022feature}.


Recently, several domain generalization methods have been developed to address this domain generalization issue~\cite{DBLP:conf/eccv/Zhang0SG22,zhou2021domain,ding2022domain,DBLP:conf/iclr/ZhouY0X21,DBLP:journals/tip/DingF18}, including augmentation-based methods, meta-learning-based methods, and domain alignment-based methods. For example, Zhang \etal~\cite{DBLP:conf/eccv/Zhang0SG22} propose a multi-view regularized meta-learning algorithm that utilizes multiple optimization trajectories to determine an appropriate direction for model updating. During testing, this method employs an ensemble scheme to generate the final prediction.
Moreover, Ding \etal~\cite{ding2022domain} aim to explicitly remove domain-specific features for domain generalization, effectively achieving domain alignment. In contrast, Xu \etal~\cite{DBLP:conf/cvpr/XuZ0W021} introduce a novel Fourier-based perspective for domain generalization. They exploit the fact that Fourier phase information contains high-level semantics and is less affected by domain shift.
Among these methods, augmentation-based methods can effectively augment the training samples, mitigating the challenge of insufficient diversity in the domain generalization task. This intuitive technique enriches the training data and addresses the limitations posed by a lack of diverse samples.

\begin{figure}
\centering
\subfigure[Art painting (A)]{
\includegraphics[width=4.13cm]{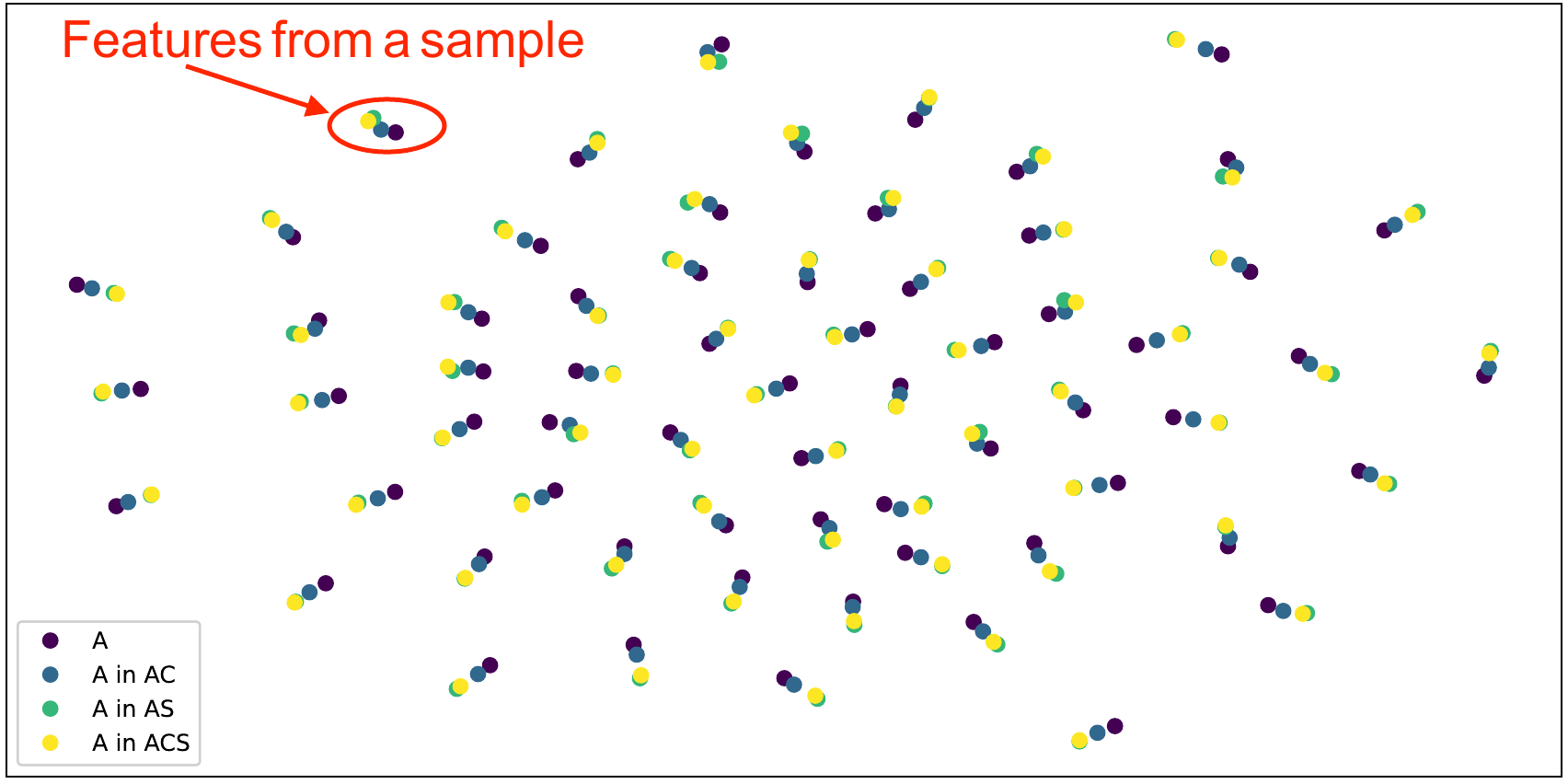}
}
\subfigure[Cartoon (C)]{
\includegraphics[width=4.12cm]{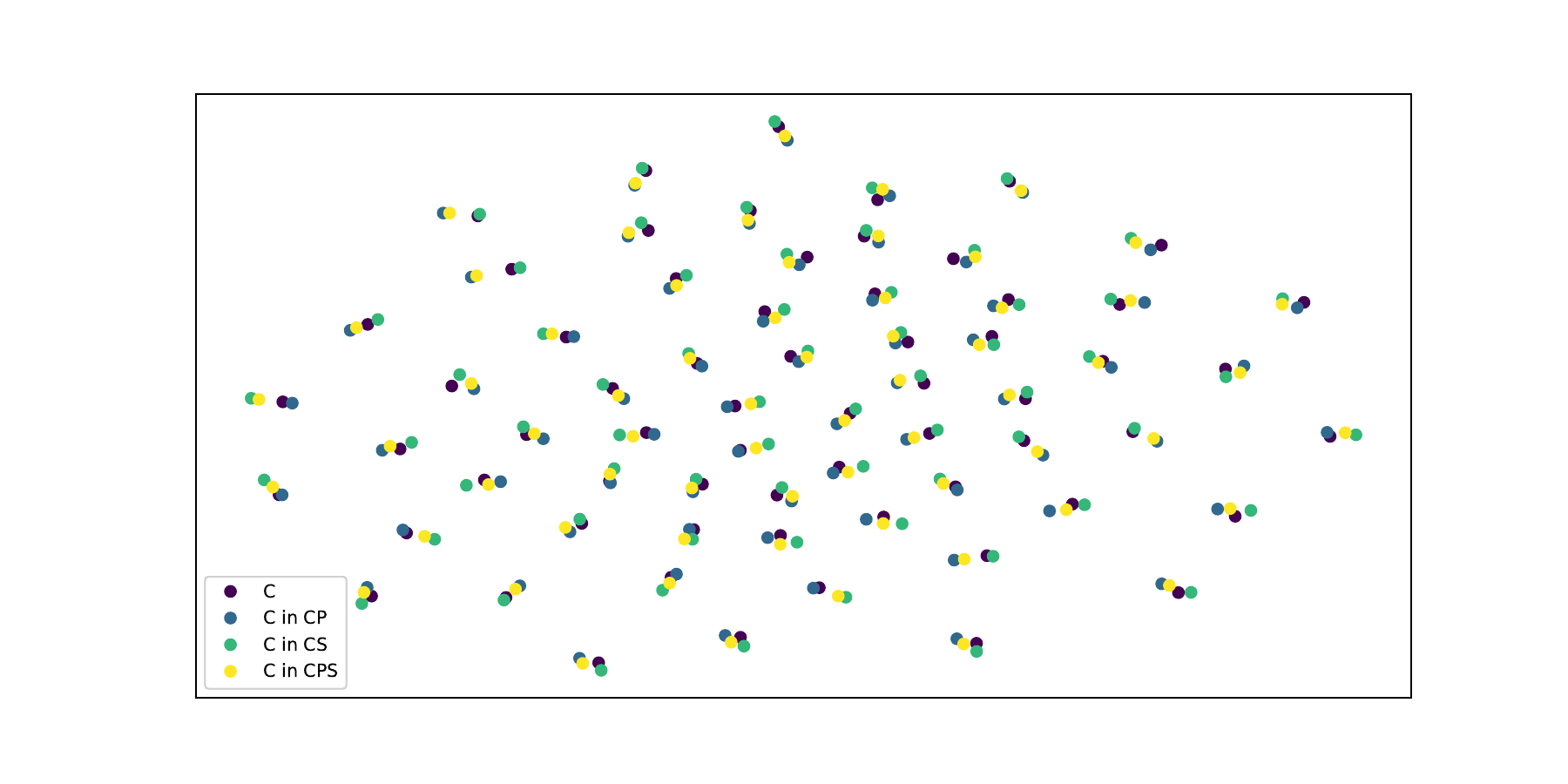}
}
\caption{Visualization of image's features by t-SNE~\cite{van2008visualizing}. 
In this figure, ``A'', ``C'', ``S'', and ``P'' represent images from the Art painting, Cartoon, Sketch, and Photo domains, respectively. For feature extraction, we utilize a ResNet-18~\cite{DBLP:conf/cvpr/HeZRS16} model pre-trained on the ImageNet dataset~\cite{DBLP:conf/cvpr/DengDSLL009}. It is important to note that we dynamically adjust the statistics ($\mu$ and $\sigma$) in all normalization layers during the feature extraction process. Each image is placed in different batches with different domains. For example, ``A in ACS'' indicates that images from the Art painting domain are combined with images from the Cartoon and Sketch domains for normalization during the test stage. As observed, this process perturbs the original feature representation.
}
\label{fig04}
\vspace{-15pt}
\end{figure}


In this paper, we propose a novel method for data augmentation in domain generalization, which differs from existing methods that perform augmentation at the image or feature level~\cite{DBLP:conf/cvpr/XuZ0W021,DBLP:conf/iclr/ZhouY0X21,DBLP:conf/cvpr/ZhangLLJZ22} to directly enrich image style information. Instead, we indirectly conduct data augmentation from a new perspective using batch normalization (BN). To visualize this concept, we perform an experiment by combining different domains, as depicted in Fig.~\ref{fig04}. During this experiment, we keep all model parameters fixed, except for the statistics in the normalization layers. Specifically, the statistics in each batch normalization layer are computed using the current batch. As shown in Fig.~\ref{fig04}, each image can be perturbed by normalizing it with different statistics derived from batches of different domains. Therefore, we can leverage and explore this observation in the context of domain generalization to enhance the diversity of training samples.

Inspired by this observation, we propose a novel normalization-guided augmentation (NormAUG) method to enhance the model's generalization. Our method consists of two paths: the main path and the augmented (or auxiliary) path. The main path is similar to the baseline method. The augmented path includes multiple sub-paths during training, with each sub-path representing batch normalization for a single domain or a random combination of multiple domains. This strategy effectively introduces diverse information into the feature representation. Importantly, these sub-paths in the auxiliary path can vary at each iteration as we randomly select sub-paths from a batch normalization (BN) bank. All paths and sub-paths in our method are implemented using different batch normalization layers, while other parameters are shared.

Additionally, we incorporate a classifier bank for the auxiliary path, where each classifier corresponds to a batch normalization (BN) layer in the BN bank. This method provides additional diverse information to our method. Leveraging the properties of our method, we combine the results from the auxiliary path with those from the main path to further enhance the model's generalization. Extensive experiments demonstrate that our method outperforms state-of-the-art methods on various benchmark domain generalization datasets. Ablation analysis confirms the effectiveness of each module in our method. Furthermore, we analyze the effectiveness of NormAUG based on existing domain generalization theory, which reveals that our method achieves a lower generalization upper bound compared to the baseline method.

In this paper, our main contributions can be summarized as:
  \begin{itemize}
    \item We develop a novel Normalization-guided Augmentation (NormAUG) for domain generalization, which can effectively enhance the diversity of training data. 
   \item We devise the BN bank and classifier bank to implement the proposed normalization-guided augmentation. This method not only enhances data diversity but also contributes to an improved ensemble prediction.
    \item Our method achieves state-of-the-art accuracy on multiple standard benchmark datasets, demonstrating its superiority over existing methods. We also provide an ablation study and further analysis to validate the effectiveness of our proposed method. 
  \end{itemize}

The structure of this paper is outlined as follows: Section \ref{s-related} provides a literature review on relevant research. In Section \ref{s-framework}, we introduce our normalization-guided augmentation method. Section \ref{s-experiment} presents the experimental results and analysis. Finally, we conclude in Section \ref{s-conclusion}.


\section{Related work}\label{s-related}
In this section, we review the most related domain generalization methods to our method, including data augmentation, ensemble learning and other methods. The following part presents a detailed investigation.

\subsection{Data Augmentation }

Since data augmentation can effectively enhance the diversity of training data, it has been recognized as a valuable method to improve the model's generalization ability in unseen domains. In recent years, several methods have been developed from this perspective for the domain generalization task.
For instance, Huang \etal~\cite{DBLP:conf/eccv/HuangWXH20} propose a simple training heuristic called Representation Self-Challenging (RSC) that significantly improves the generalization of convolutional neural networks (CNN) to out-of-domain data. RSC iteratively challenges the dominant features activated on the training data and encourages the network to activate remaining features that are more correlated with the labels.
Another method is introduced by Xu \etal~\cite{DBLP:conf/cvpr/XuZ0W021}, who introduce a novel Fourier-based perspective for domain generalization. Their method leverages the assumption that the Fourier phase information contains high-level semantics and is not easily affected by domain shift, thus providing a robust representation for generalization across domains.
In addition, Wang \etal~\cite{DBLP:conf/iccv/WangLQHB21} develop a style-complement module to enhance the generalization power of the model. This module synthesizes images from diverse distributions that are complementary to the source domain, thereby enriching the training data and improving the model's ability to generalize to unseen domains.


Recently, there has been an increasing interest in augmentation methods based on Instance Normalization (IN), inspired by the AdaIN technique proposed by Huang \etal~\cite{DBLP:conf/iccv/HuangB17}. These methods aim to enhance the diversity and generalization of models through instance-level normalization.
For example, Zhang \etal~\cite{DBLP:conf/cvpr/ZhangLLJZ22} propose a method called Exact Feature Distribution Matching (EFDM) that matches the empirical cumulative distribution functions of image features. This is achieved by applying exact histogram matching in the image feature space, enabling precise feature distribution alignment.
Another method, introduced by Kang \etal~\cite{DBLP:conf/cvpr/KangLKK22}, involves synthesizing novel styles continuously during training. They manage multiple queues to store observed styles and synthesize novel styles with distinct distributions compared to the styles in the queues. This method aims to enrich the style diversity and improve generalization.
Additionally, MixStyle~\cite{DBLP:conf/iclr/ZhouY0X21} explores a technique that probabilistically mixes instance-level feature statistics of training samples across source domains. By blending feature statistics, MixStyle encourages the model to learn more robust and domain-invariant representations, enhancing generalization performance.

Unlike the methods mentioned above, our method focuses on data augmentation using Batch Normalization (BN). By leveraging BN, we aim to effectively explore the diversity present in the training data. This allows us to enhance the generalization ability of our model by introducing variations in the normalization process.

\subsection{Ensemble Learning}

In the domain of domain generalization, ensemble learning has been widely utilized to improve prediction accuracy by leveraging multiple experts during the test process. For instance, Niu \etal~\cite{DBLP:conf/cvpr/NiuLX15} extend the multi-class SVM formulation to train a classifier for each class and latent domain, effectively integrating multiple classifiers to enhance generalization capability. Seo \etal~\cite{DBLP:conf/eccv/SeoSKKHH20} propose a simple yet effective multi-source domain generalization technique based on deep neural networks, incorporating optimized normalization layers specific to individual domains. Zhou \etal~\cite{zhou2021domain} introduce the domain adaptive ensemble learning (DAEL) framework, comprising a shared CNN feature extractor and multiple classifier heads trained to specialize in different source domains. Each classifier acts as an expert for its own domain and a non-expert for others. Segu \etal~\cite{DBLP:journals/pr/SeguTT23} train domain-dependent representations using ad-hoc batch normalization layers to collect independent domain statistics, enabling mapping of domains in a shared latent space. At test time, samples from an unknown domain are projected into this space to infer domain properties. Besides, Zhang \etal~\cite{DBLP:conf/eccv/Zhang0SG22} employ a multi-view meta-learning scheme in the training stage and utilize an ensemble scheme by producing multiple test images for a sample to generate the final fused prediction.

In our work, our primary focus is on data augmentation rather than designing an ensemble scheme. It is important to note that the ensemble scheme serves as a supplementary component in the test stage of our method. Furthermore, unlike the method by Zhang \etal~\cite{DBLP:conf/eccv/Zhang0SG22}, our method does not require augmenting the test images during testing.

\subsection{Other Methods}
Besides the methods mentioned above, domain alignment or learning domain-invariant features is also crucial in domain generalization (DG)~\cite{DBLP:conf/aaai/LiGTLT18,DBLP:conf/aaai/MatsuuraH20,DBLP:conf/eccv/ChattopadhyayBH20,DBLP:conf/eccv/MengLCYSWZSXP22,DBLP:conf/eccv/LeeKK22}. For example, Li \etal~\cite{DBLP:conf/aaai/LiGTLT18} propose learning features with domain-invariant class conditional distributions. 
Furthermore, meta-learning methods simulate training/test domain shift during training by synthesizing virtual test domains within each mini-batch \cite{DBLP:conf/icml/FinnAL17,DBLP:conf/nips/BalajiSC18,DBLP:conf/aaai/LiYSH18,DBLP:conf/nips/DouCKG19,lv2023improving}. For example, Zhang \etal~\cite{DBLP:conf/eccv/Zhang0SG22} develop a multi-view regularized meta-learning algorithm employing multiple optimization trajectories. These methods aim to enhance model generalization by aligning features across all domains or using meta-learning techniques.

\section{The proposed method}\label{s-framework}

\begin{figure}[t]
\centering
\includegraphics[width=8.7cm]{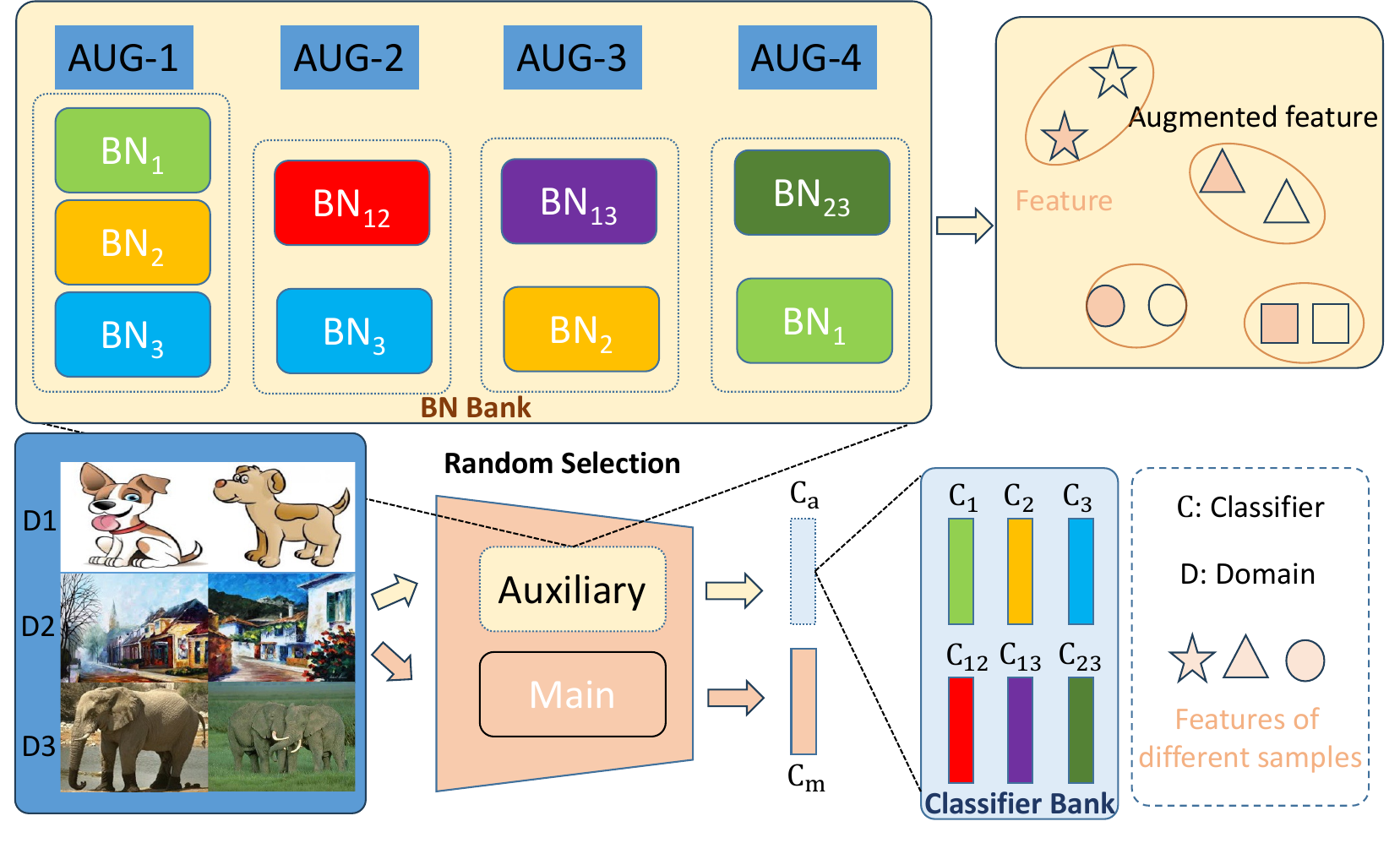}
\caption{The pipeline of our method in the training stage. 
In our training setup, we assume that there are three domains in the training set. The network architecture consists of two paths: the main path and the auxiliary path. Specifically, the data augmentation is performed through the auxiliary path. During each iteration, we randomly select a set of batch normalization (BN) layers from the BN bank (\eg, AUG-2 in the figure) and the corresponding classifier from the classifier bank to train the model. The main path and the auxiliary path share all parameters except for the BN layers. It is worth noting that BN layers with the same color in the BN bank share parameters, and all normalization layers in the auxiliary path are replaced by a BN bank. The corresponding algorithm is shown in Alg.~\ref{al01}.
}
\label{fig05}
\vspace{-15pt}
\end{figure}

In this paper, we are inspired by the observation depicted in Fig.~\ref{fig04}, where the normalization of images from different domains introduces variations in their features. Motivated by this observation, we propose a novel method called normalization-guided augmentation (NormAUG) for domain generalization, as illustrated in Fig.~\ref{fig05}. Our method comprises two paths: the main path and the auxiliary path, which performs data augmentation using a batch normalization bank. Meanwhile, we also employ a classifier bank to better train our model. In the following sections, we will provide a detailed explanation of the background and our proposed method.

\subsection{Background}
Here, we will review the conventional batch normalization (BN)~\cite{DBLP:conf/icml/IoffeS15}. First, we define feature maps of an image $f_k\in \mathbb{R}^{C \times H \times W}$, where $C$ are the number of channels, and $H$ and $W$ are the height and the width of feature maps. In general, BN leverages a global statistics of a batch to normalize all samples at each iteration, which can be defined as:
\begin{equation}
  \begin{aligned}
  &{\rm BN}(f_k)= \gamma \frac{f_k-\mu}{\sigma}+\beta,
  \end{aligned}
  \label{eq01}
  \end{equation}
  where  $\gamma,\beta \in \mathbb{R}^{C}$ are learnable affine transformation parameters, and $\mu, \sigma \in \mathbb{R}^{C}$ (\ie, $\mu=[\mu_1, \cdots,\mu_C]$ and  $\sigma=[\sigma_1, \cdots,\sigma_C]$) represent the channel-wise mean and standard deviation (\ie, statistics) of BN for feature maps. For statistics of the $i$-th channel are presented as:
  \begin{equation}
  \mu_i=\frac{1}{|\mathcal{B}|HW}\sum_{n\in\mathcal{B}}\sum_{h=1}^{H}\sum_{w=1}^{W}f[n,i, h,w],
    \label{eq02}
  \end{equation}
  \begin{equation}
  \sigma_i=\sqrt{\frac{1}{|\mathcal{B}|HW}\sum_{n\in\mathcal{B}}\sum_{h=1}^{H}\sum_{w=1}^{W}(f[n,i,h,w]-\mu_i)^2 + \epsilon},
  \label{eq03}
  \end{equation}
  where $\mathcal{B}$ is a batch of samples, and $|\mathcal{B}|$ is the batch size. Besides, $\epsilon$ is a constant for numerical stability.

According to Eqs.~\ref{eq01}~\ref{eq02}~\ref{eq03}, we can see that the normalized feature maps of an image are related to the statistics of a batch, meanwhile the statistics are decided by all samples of a batch. Therefore, if we randomly select one or multiple domain(s) to form a batch to perform the normalization, the diverse information can be introduced to implement data augmentation. We can also find this observation in Fig.~\ref{fig04}.

\subsection{NormAUG}

Based on the analysis above, we propose a novel data augmentation scheme called NormAUG, which leverages the batch normalization (BN) perspectively. Our model consists of two paths during the training stage: the main path and the auxiliary path. The main path serves as the baseline model, using ResNet-18 or ResNet-50~\cite{DBLP:conf/cvpr/HeZRS16} in our implementation. The auxiliary path generates augmented information through the normalization-guided argumentation.

To be more specific, we randomly select an equal number of images from $N$ domains to create a batch $\mathcal{B}=[\mathcal{B}_1;\cdots;\mathcal{B}_N]$, where $\mathcal{B}_i$ denotes the samples from the $i$-th domain in a batch. This batch is fed into both the main path and the auxiliary path. In the auxiliary path, we generate diverse information by randomly combining images from different domains and applying normalization. For instance, if we have 3 source domains, we can create four types of sub-batch combinations: $\{\mathcal{B}_1, \mathcal{B}_2, \mathcal{B}_3\}$, $\{\mathcal{B}_{12}, \mathcal{B}_3\}$, $\{\mathcal{B}_1, \mathcal{B}_{23}\}$, and $\{\mathcal{B}_2, \mathcal{B}_{13}\}$, for the auxiliary path. Here, $\mathcal{B}_{12}$ is the sub-batch consisting of the samples from the first and second domains in a batch. 

During the training process, our method employs a BN bank for each normalization layer in the auxiliary path, which consists of the corresponding four BN combinations: $\{\mathrm{BN}_1, \mathrm{BN}_2, \mathrm{BN}_3\}$, $\{\mathrm{BN}_{12}, \mathrm{BN}_3\}$, $\{\mathrm{BN}_1, \mathrm{BN}_{23}\}$, and $\{\mathrm{BN}_2, \mathrm{BN}_{13}\}$, as shown in Fig.~\ref{fig05}. When we generate the sub-batch set $\{\mathcal{B}_{12}, \mathcal{B}_3\}$, we feed it into the sub-paths using $\{\mathrm{BN}_{12}, \mathrm{BN}_3\}$ in each normalization layer. Besides, we feed $\mathcal{B}_{123}$ into the main path. It is important to note that the main path and the auxiliary path share the same parameters, except for the BN layers. Therefore, our method does not introduce a large of extra parameters.



\begin{figure}
\centering
\includegraphics[width=8cm]{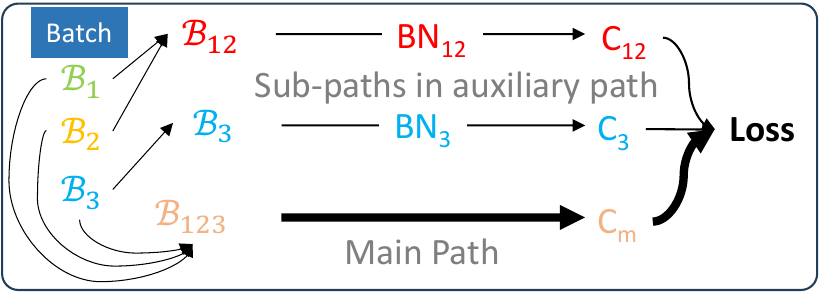}
\caption{The forward process of our method in the training stage. Here, we assume there are 3 source domains. This figure denotes that we randomly select the ``AUG-2'' in Fig.~\ref{fig05} from the BN bank at an iteration.
}
\label{fig09}
\vspace{-15pt}
\end{figure}
In this paper, we adopt multiple classifiers to train our model. The main path consists of an independent classifier, while in the augmented path, all BNs in the normalization bank have their respective independent classifiers. For instance, in the case of 3 source domains, we have the classifier $\mathrm{C}_m$ for the main path, and the classifier set $\{\mathrm{C}_1, \mathrm{C}_2, \mathrm{C}_3, \mathrm{C}_{12}, \mathrm{C}_{13}, \mathrm{C}_{23}\}$ for the augmented path. Each classifier corresponds to the BN set used in the respective sub-paths. The forward process of our method is illustrated in Fig.~\ref{fig09}. The overall training loss can be defined as follows:

\begin{equation}
\begin{aligned}
 \mathcal{L}= &\sum_{x \in \mathcal{B}}\sum_{c=1}^{C}-\log(P(c|x))\cdot\mathbf{I}(x\in c) \\
&+\frac{1}{K}\sum_{k=1}^{K}\sum_{x \in \mathcal{B}_{k}}\sum_{c=1}^{C}-\log(P_k(c|x))\cdot\mathbf{I}(x\in c),
\end{aligned}
  \label{eq04}
  \end{equation}
where $K$ is the number of random sub-batches. $\mathbf{I}(x\in c)$ is equal to 1 if $x\in c$, otherwise 0. The detailed training process is described in Alg.~\ref{al01}.

 \subsection{Discussion} 
\underline{The scheme of the combination.}
Generally, for $N$ domains, there are $1+C_N^2+C_N^3\cdots+C_N^{N-1}$ types of sub-batch combinations. Although adding BN layers does not bring a large of extra parameters, our method requires using the independent classifier for each BN, which results in the large classifier bank in the training process. Therefore, when there are N domains, we produce $N+1$ types of sub-batch combinations. For example, we generate 5 types of sub-batch combinations: $\{\mathcal{B}_1, \mathcal{B}_2, \mathcal{B}_3, \mathcal{B}_4\}$, $\{\mathcal{B}_{123}, \mathcal{B}_4\}$, $\{\mathcal{B}_{124}, \mathcal{B}_{3}\}$, $\{\mathcal{B}_{134}, \mathcal{B}_{2}\}$,  and $\{\mathcal{B}_{234}, \mathcal{B}_{1}\}$ for 4 domains.

\begin{figure}[t]
\centering
\includegraphics[width=8.5cm]{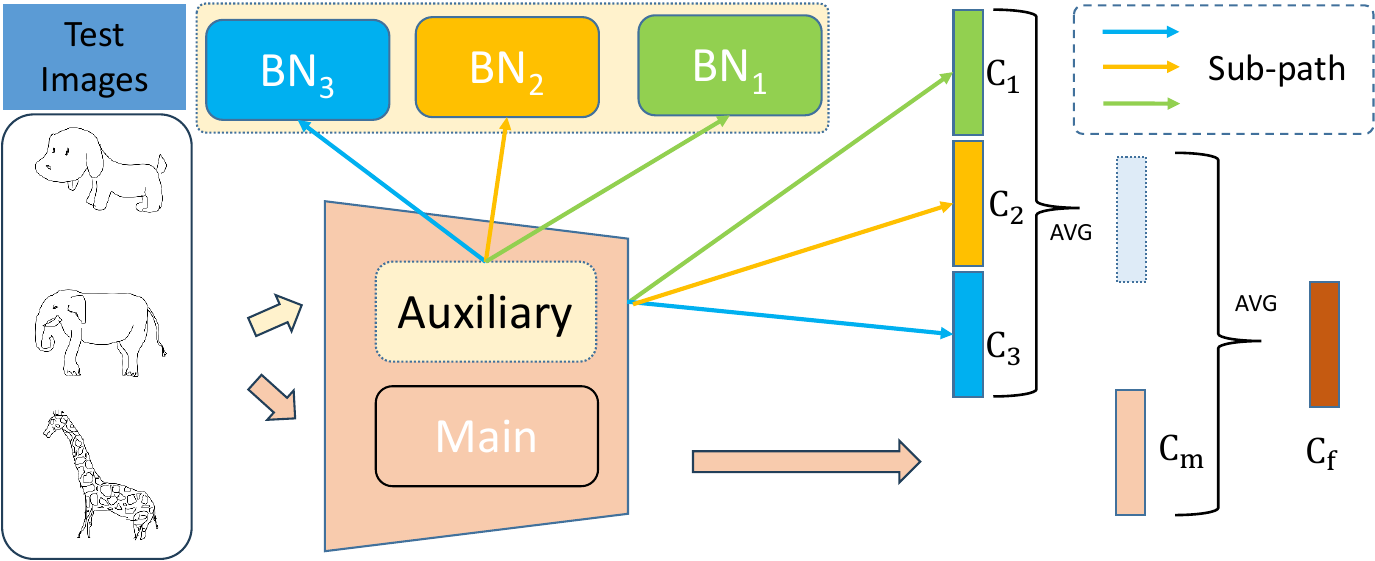}
\caption{The pipeline of our method in the test stage. In the test stage, we fuse the results from sub-paths (\ie, ``AUG-1'' in Fig.~\ref{fig05}) to further improve the accuracy of the $\mathrm{C}_m$. In this figure, ``AVG'' denotes the average operation.
}
\label{fig06}
\end{figure}

\underline{The ensemble prediction.}
In addition, while our method aims to improve the accuracy of the main path through the auxiliary path, we can further enhance the prediction by employing a fusion scheme. By considering the diversity among the sub-paths in the auxiliary path, we select only the single-domain path to obtain an averaged result for the augmented path. The final prediction is then obtained by averaging the results from both the main path and the augmented path. This fusion process is illustrated in Fig.~\ref{fig06}, where the combined result provides an improved prediction. The detailed test process is described in Alg.~\ref{al02}.

\underline{The other tricks.}
Instance normalization (IN) is known for its ability to remove domain-specific (style) information and improve the generalization of models. Therefore, in the main path of our method, we combine instance normalization (IN) with batch normalization (BN) to further enhance the model's generalization. Specifically, in this paper, we utilize optimized normalization~\cite{DBLP:conf/eccv/SeoSKKHH20} in the main path of our model.

\underline{The benefits of the classifier bank.}
The advantages of using the classifier bank in the auxiliary path can be summarized as follows:
1) Enhanced Data Diversity: By using a single classifier for each BN in the auxiliary path, we can effectively explore diverse information, as each BN is free to adapt to the specific domain characteristics. This flexibility allows us to avoid any underlying constraints during training and improve the model's generalization.
2) Improved Ensemble Prediction: The independent classifiers can also be leveraged for the fused prediction in the test stage. This ensemble scheme further enhances the overall performance and will be validated in our experiments.
By employing the classifier bank in the auxiliary path, our method can capitalize on these benefits, leading to superior results in domain generalization tasks.

\underline{Comparison with the most relevant methods.}
We compare our NormAUG method with two closely related methods, namely MixStyle~\cite{DBLP:conf/iclr/ZhouY0X21}, DSON~\cite{DBLP:conf/eccv/SeoSKKHH20}, and BEN~\cite{DBLP:journals/pr/SeguTT23}. Compared to MixStyle, there are three key differences:
1) The underlying technique is different. Our NormAUG method is based on batch normalization, while MixStyle is based on instance normalization.
2) Our method employs a BN bank, which allows for the generation of more diverse samples, whereas MixStyle always mixes the styles of two images.
3) Our NormAUG method can produce an ensemble result in the test stage, thanks to its inherent properties, while MixStyle does not have this capability.
In comparison to DSON, there are two main differences:
1) While DSON has independent paths for each domain, our NormAUG method includes multiple sub-paths in the auxiliary path, such as combinations of different domains.
2) The goals of the two methods differ. Our method has a main path, and our objective is to utilize the auxiliary path to enhance the generalization of the main path. On the other hand, DSON aims to learn multiple experts to improve prediction performance in unseen domains.
It is worth noting that our NormAUG method, without the ensemble prediction in the test stage, can also achieve excellent performance, as demonstrated in Table~\ref{tab05}. This is attributed to the diverse features obtained from the auxiliary path during training.
In addition, BEN~\cite{DBLP:journals/pr/SeguTT23} is trained by independently evaluating domains to obtain feature embeddings of source domains. It then calculates the distance function between unknown domain samples and the source domain to linearly weight the representation of unknown domain samples during testing. Our method employs the BN bank and classifier bank for indirectly enhancing the diverse information in the training stage, and fuses different sub-paths to achieve the final prediction in the testing stage.

\begin{algorithm}[t]\label{al01}
    \SetAlgoLined 
	\caption{\small{The training process of our NormAUG}}
	\KwIn{Training samples $\mathrm{X}_{tr}$ and labels $\mathrm{Y}$.}
	\KwOut{The trained model ($\theta$).}
	 $\theta \leftarrow$ Initialize by ResNet pre-trained on ImageNet. \\ 
	 \tcp{\bl{\scriptsize{The number of epochs is $T$.}}} 
	\For{epoch $\in [1,...,T]$}{
		\tcp{\bl{\scriptsize{The number of iterations in each epoch is $N$.}}}
		\For{iteration $\in [1,...,N]$}{
		 Randomly select a combination of the normalization and classifier for the auxiliary path as shown in Fig.~\ref{fig05}.\\
		 Feed these input images into the main path and the auxiliary path, respectively.\\
		 Compute the whole loss as Eq.~\ref{eq04}.\\
		 Update the model parameter $\theta$.
		}
	}
	return 
\end{algorithm}

\begin{algorithm}[t]\label{al02}
    \SetAlgoLined 
	\caption{\small{The test process of our NormAUG}}
	\KwIn{Test samples $\mathrm{X}_{te}$.}
	\KwOut{The predicted result.}
	 $\theta \leftarrow$ Initialize by the trained model $\theta$. \\ 
		\tcp{\bl{\scriptsize{The number of iterations in the test set is $N_{te}$.}}}
		\For{iteration $\in [1,...,N_{te}]$}{
		 Feed the test images into the main path and the auxiliary path.\\
		 Obtain the result from the main path ($p_{m}$), and the result set from the auxiliary path ($P_{a}$).\\
		 Compute the average result of sub-paths in the auxiliary path as $\bar{p}_a$.\\
		 Compute the average of $p_{m}$ and $\bar{p}_a$ as $p_{f}$.\\
		}
	return 
\end{algorithm}

\underline{Explanation for NormAUG via existing theory.}
In this section, we use the domain generalization error bound~\cite{albuquerque2019generalizing} to demonstrate the effectiveness of our method. Firstly, we review the domain generalization error bound, and then we analyze our method based on it.

\textbf{Theorem 1}~\cite{albuquerque2019generalizing,DBLP:conf/ijcai/0001LLOQ21} (Domain generalization error bound): Let $\gamma := min_{\pi \in \bigtriangleup_{M}} d_{\mathcal{H}}(\mathcal{P}_{X}^t, \sum_{i=1}^{M}\pi_{i}\mathcal{P}_{X}^i)$\footnote{$M$ is the number of source domains.} with minimizer $\pi^*$ being the distance of $P_{X}^t$ from the convex hull $\Lambda$. Let $P_{X}^* := \sum_{i=1}^{M}\pi_{i}^{*}P_{X}^{i}$ be the best approximator within $\Lambda$. Let $\rho := \sup_{\mathcal{P}_{X}^{'}, \mathcal{P}_{X}^{''}\in \Lambda}d_{\mathcal{H}}(\mathcal{P}_{X}^{'}, \mathcal{P}_{X}^{''})$ be the diameter of $\Lambda$. Then it holds that

\begin{equation}
\begin{aligned}
\epsilon^{t}(h)\leqslant \sum_{i=1}^{M}\pi_{i}^{*}\epsilon^{i}(h)+\frac{\gamma+\rho}{2}+\lambda_{\mathcal{H}}(\mathcal{P}_X^t, \mathcal{P}_X^*)),
\end{aligned}
\label{eq22}
\end{equation}
where $\lambda_{\mathcal{H}}(\mathcal{P}_X^t, \mathcal{P}_X^*)$ is the ideal joint risk across the target domain and the training domain ($P_X^*$) with the most similar distribution to the target domain.
In Theorem 1, the last item can be treated as a constant because it represents the ideal joint risk across the target domain and the training domain ($P_X^*$) with the most similar distribution to the target domain. Besides, 
the first item is the empirical risk error of all source domains. Most of existing methods use the cross-entropy loss to train the model, thus it can also be viewed as a constant in the upper bound. Therefore, we primarily focus on analyzing the second item (which consist of $\gamma$ and $\rho$). 
\begin{itemize}
 \item $\gamma$ represents the discrepancy between the combination of all training domains and the target domain. In the domain generalization setting, if the test domain is far from the training domain in terms of distribution, the model's generalization may be poor for all test samples. Particularly, our method utilizes the normalization based augmentation to enrich the diversity, resulting in obtaining the domain-invariant feature. We compute the divergence between the source and target domains, as shown in Tab.~\ref{tab07} of the experimental section. From this table, we observe that the model with our augmentation generates a smaller domain gap between source and target domains than  the model without our augmentation. Therefore, introducing our NormAUG can be beneficial in reducing the influence of the domain-shift between training and test sets and effectively mitigating the aforementioned risk.
 \item $\rho$ indicates the maximum distance between different source domains. In our method, the normalization based augmentation scheme preserves semantic information while not introducing additional information. This indicates that \textit{indirectly} generating diverse style information in our method does not create a large domain gap between training samples. Additionally, we can also observe this fact in Fig.~\ref{fig04}.
 In summary, our method has the advantage of reducing the generalization error bound from the second item in Eq.~\ref{eq22}.
\end{itemize}
\section{Experiments}\label{s-experiment}
\renewcommand{\cmidrulesep}{0mm} 
\setlength{\aboverulesep}{0mm} 
\setlength{\belowrulesep}{0mm} 
\setlength{\abovetopsep}{0cm}  
\setlength{\belowbottomsep}{0cm}

 In this section, we begin by presenting the experimental datasets and configurations in Section~\ref{sec:EXP-DS}. Following that, we evaluate our proposed method against the current state-of-the-art domain generalization techniques in Section~\ref{sec:EXP-CUA}. To verify the impact of different modules in our framework, we carry out ablation studies in Section~\ref{sec:EXP-SS}. Finally, we delve deeper into the properties of our method in Section~\ref{sec:EXP-FA}.
\subsection{Datasets and Experimental Settings}\label{sec:EXP-DS}
\begin{figure}[t]
\centering
\includegraphics[width=8.5cm]{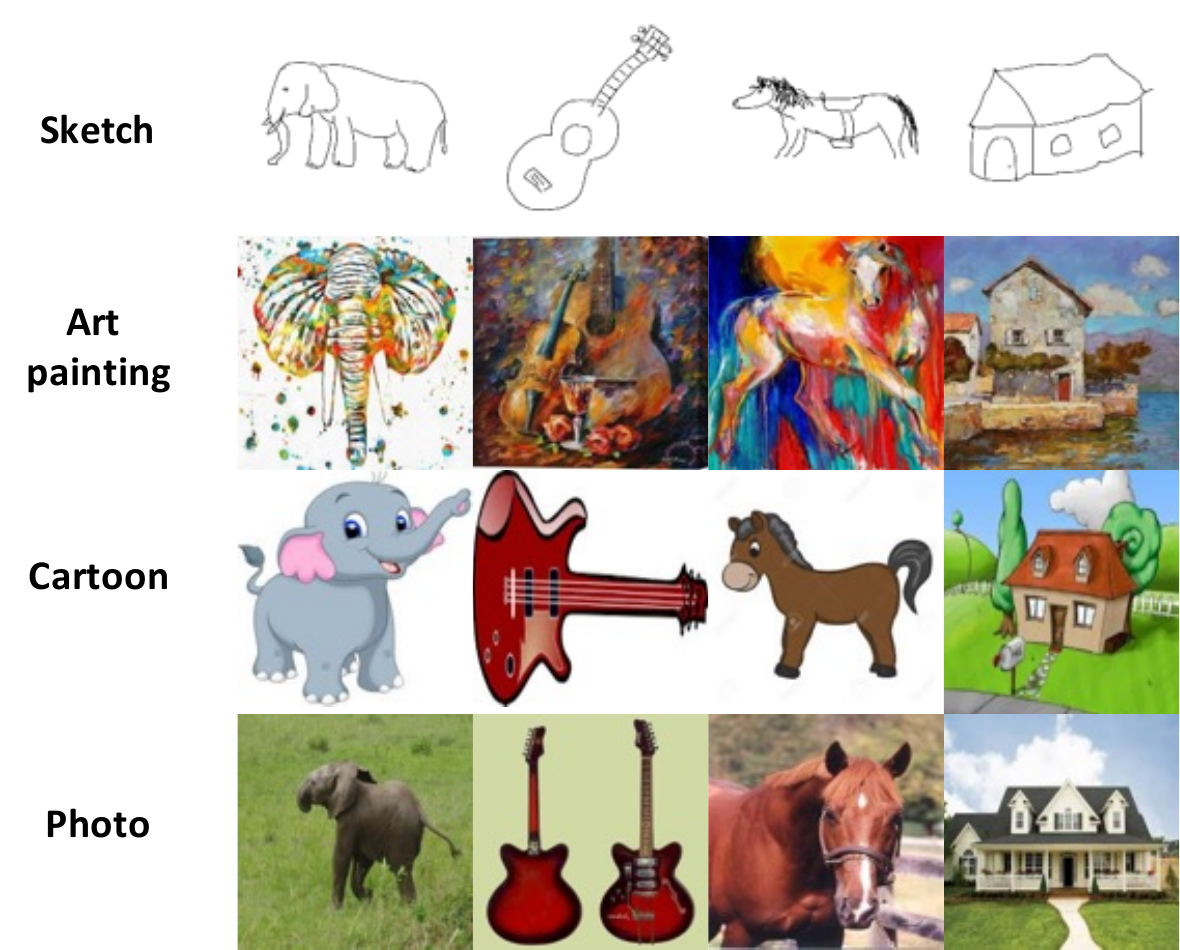}
\caption{Examples from PACS. We show some images from Sketch, Art painting, Cartoon, and Photo. As seen in this figure, the difference is obvious for these images with the same class from different domains.
}
\label{fig08}
\end{figure}
\begin{figure}
\centering
\subfigure[Elephant]{
\includegraphics[width=4.0cm]{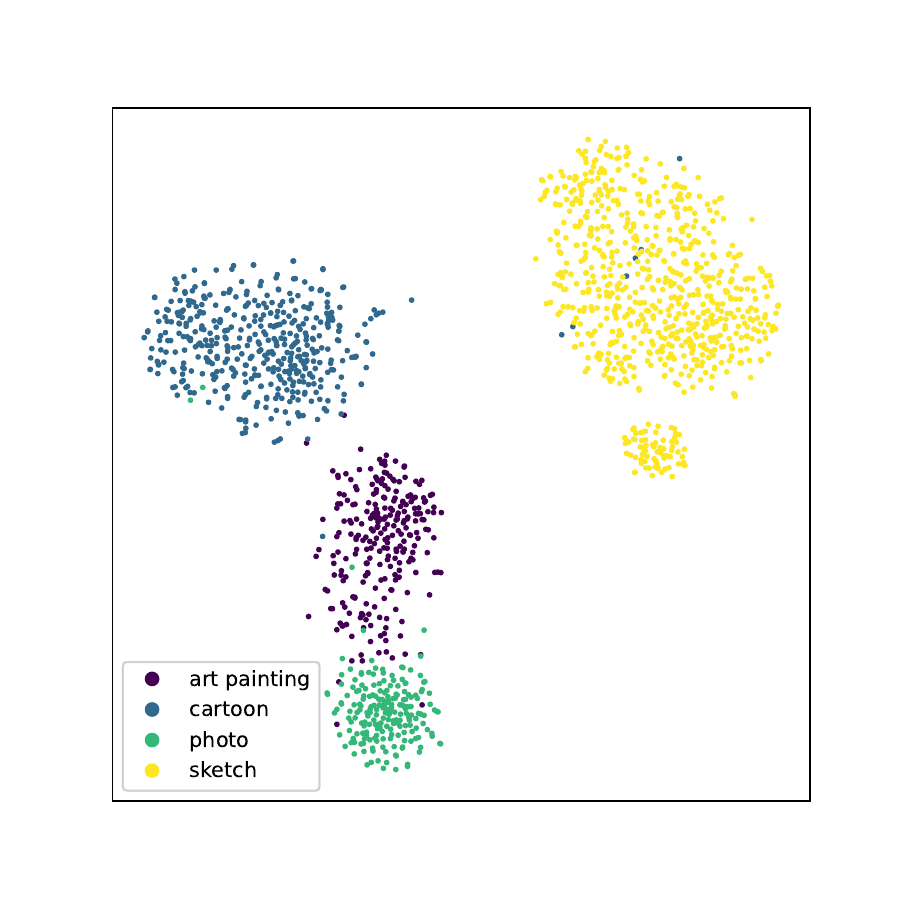}
}
\subfigure[Guitar]{
\includegraphics[width=4.0cm]{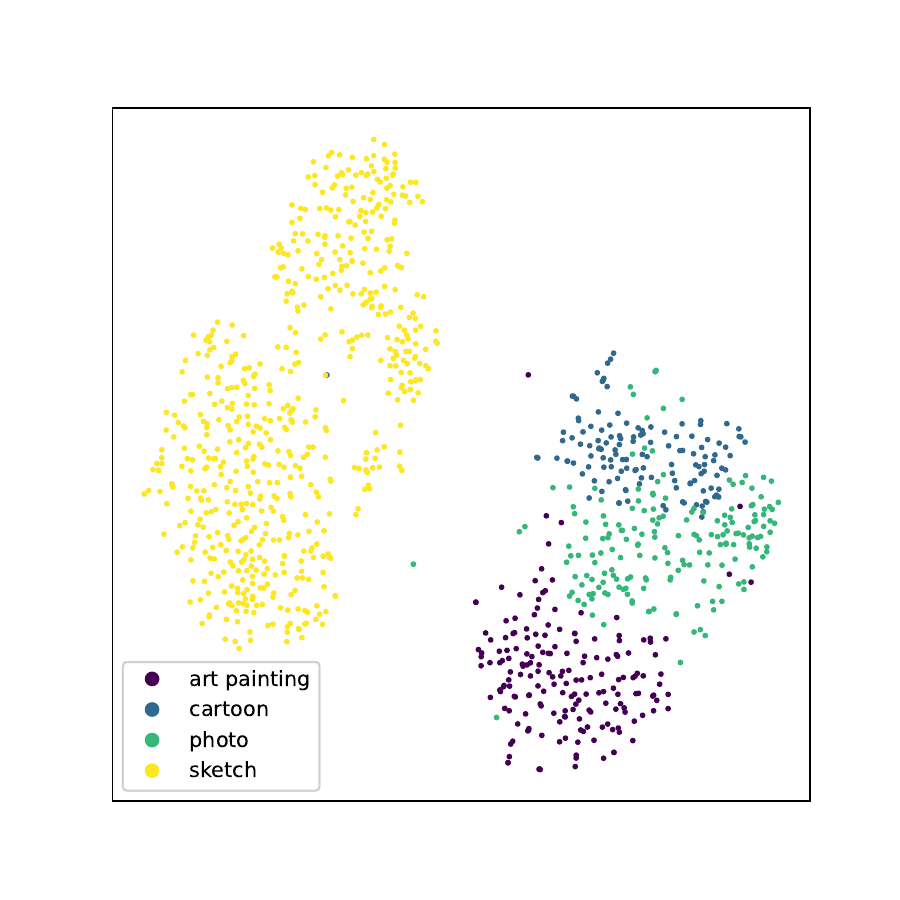}
}
\subfigure[Horse]{
\includegraphics[width=4.0cm]{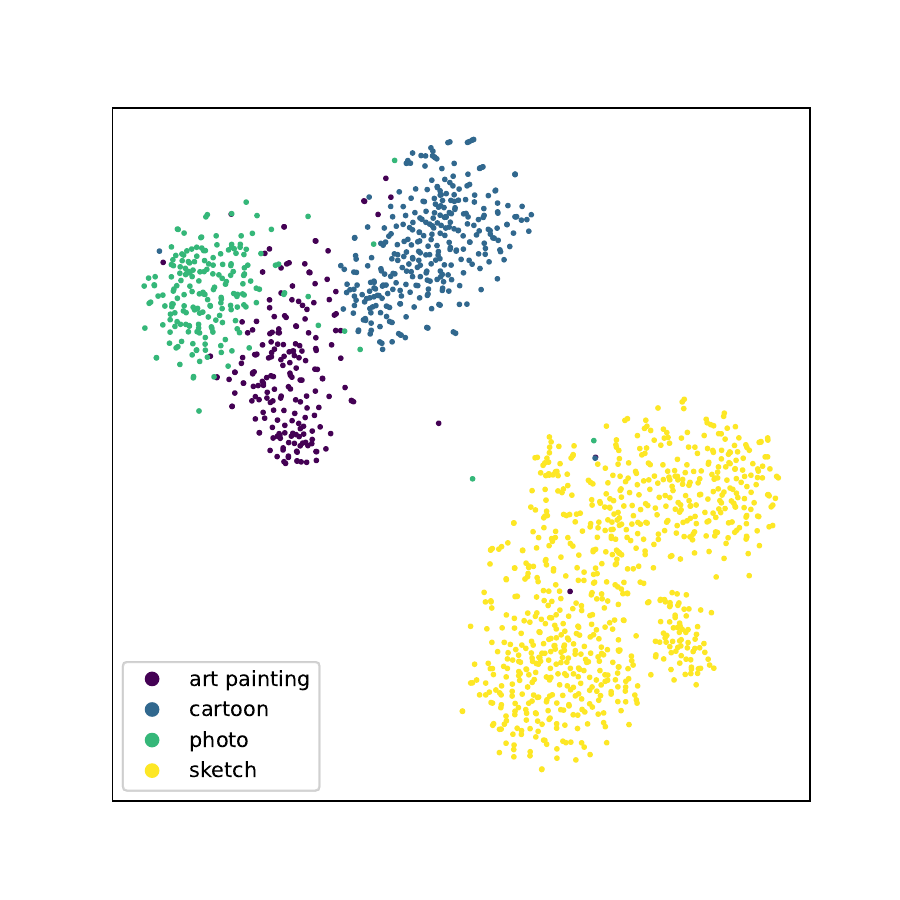}
}
\subfigure[House]{
\includegraphics[width=4.0cm]{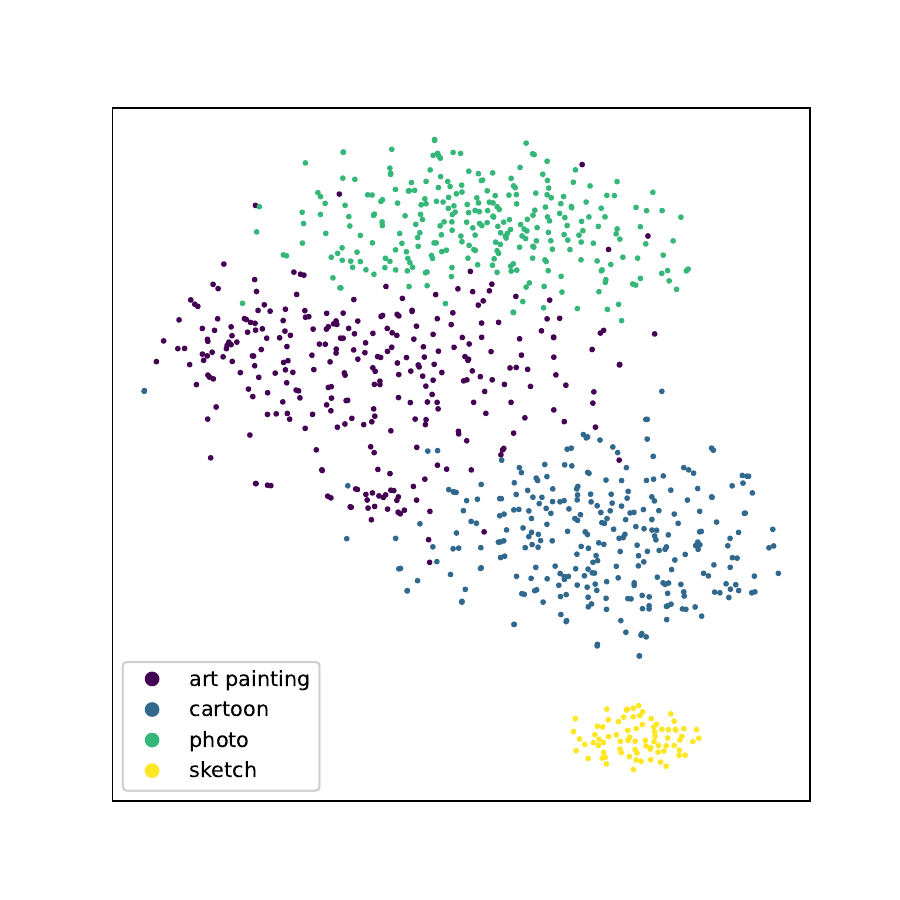}
}
\caption{Visualization of image's features by t-SNE~\cite{van2008visualizing}. In this figure, we extract the image's features using ResNet-18~\cite{DBLP:conf/cvpr/HeZRS16} pre-trained on ImageNet to show the domain gap from the feature representation view. As observed in this figure, these images from the same class are in different positions, which can obviously show the domain's difference as Fig.~\ref{fig08}.}
\label{fig07}
  \vspace{-15pt}
\end{figure}
\subsubsection{Datasets} 
In this paper, we conduct the experiments to validate the effectiveness of our method on five benchmark DG datasets as follows:
  \begin{itemize}
      \item \textbf{PACS} \cite{Li2017DeeperBA} consists of four different domains: Photo, Art painting, Cartoon and Sketch. It contains 9,991 images with 7 object categories in total, including Photo (1,670 images), Art (2,048 images), Cartoon (2,344 images), and Sketch (3,929 images).
      \item \textbf{Office-Home} \cite{venkateswara2017deep} contains 15,588 images of 65 categories of office and home objects. It has four different domains namely Art (2,427 images), Clipart (4,365 images), Product (4,439 images) and Real World (4,357 images), which is originally introduced for UDA but is also applicable in the DG setting.
      \item \textbf{DomainNet} ~\cite{DBLP:conf/iccv/PengBXHSW19}
      is an extensive domain generalization dataset, comprising 596,010 images distributed across 345 categories from 6 distinct domains: Clipart (48,837 images), Infograph (53,201 images), Painting (75,759 images), Quickdraw (172,500 images), Real (175,327 images), and Sketch (70,386 images).
      \item \textbf{mini-DomainNet} \cite{zhou2021domain} takes a subset of DomainNet~\cite{DBLP:conf/iccv/PengBXHSW19}. mini-DomainNet includes four domains and 126 classes. As a result, mini-DomainNet contains 18,703 images of Clipart, 31,202 images of Painting, 65,609 images of Real and 24,492 images of Sketch.
\item   \textbf{DigitsDG}~\cite{zhou2020deep} is a digit recognition benchmark consisting of four
classical datasets MNIST~\cite{lecun1998gradient}, MNIST-M~\cite{ganin2015unsupervised}, SVHN~\cite{netzer2011reading},
SYN~\cite{ganin2015unsupervised}. The four datasets mainly differ in font style,
background and image quality. We use the original trainvalidation split in~\cite{zhou2020deep} with 600 images per class per dataset.

  \end{itemize}
We show some examples from four different domains on PACS Fig.~\ref{fig08}. As seen, there is an obvious difference among different domains. Besides, we also visualize the features of four categories on PACS by t-SNE~\cite{van2008visualizing}, as illustrated in Fig.~\ref{fig07}. In this figure, different colors denote different domains. We observe that different domains appear in different spaces, validating that there exists the domain shift in the training set.

\subsubsection{Implementation Details} 
In this study, we utilize ResNet-18 ~\cite{DBLP:conf/cvpr/HeZRS16} and ResNet-50 ~\cite{DBLP:conf/cvpr/HeZRS16}  models pretrained on the ImageNet ~\cite{DBLP:conf/cvpr/DengDSLL009}  dataset as the backbone for our framework. All images are resized to dimensions of 224 $\times$ 224. We randomly sample 16 images from each domain to form a batch of data for input to our network. During training, we apply data augmentations including horizontal flipping, random cropping, color jittering.
We use the SGD optimizer for both the classifier and backbone networks. By default, we set the initial learning rate ($lr_c$) for the classifier as 0.01, and the initial learning rate ($lr_b$) for the backbone to 0.003. Especially, for the Office-Home dataset, we use $lr_c$ and $lr_b$ as 0.005 and 0.001, respectively. We follow the standard data splits and adopt the leave-one-domain-out evaluation protocol as used in ~\cite{DBLP:conf/cvpr/CarlucciDBCT19}. For evaluating the performance on the target domain, we select the model from the final training epoch. The accuracy on the target domain is reported and averaged over three runs. We employ the same settings for  experiments on all datasets.

\subsection{Comparison with State-of-the-art Methods}\label{sec:EXP-CUA}

In this section, we compare our NormAUG method with several state-of-the-art (SOTA) methods on four benchmark datasets: PACS, Office-Home, mini-DomainNet, Digits-DG, and DomainNet. The experimental results are reported in Tabs.~\ref{tab01}, \ref{tab02}, \ref{tab03}, \ref{tab04}, and \ref{tab11}. In the following part, we will give the detailed analysis.

\textbf{Results on PACS.} 
We compare our NormAUG method with several augmentation-based methods, including EFDMix~\cite{DBLP:conf/cvpr/ZhangLLJZ22}, FACT~\cite{DBLP:conf/cvpr/XuZ0W021}, RSC~\cite{DBLP:conf/eccv/HuangWXH20}, and STNP~\cite{DBLP:conf/cvpr/KangLKK22}. The experimental results are shown in Tab.~\ref{tab01}. As observed, our NormAUG consistently outperforms these methods on both ResNet-18 and ResNet-50. Additionally, we compare our method with ensemble methods such as DAEL~\cite{zhou2021domain} and DSON~\cite{DBLP:conf/eccv/SeoSKKHH20}, demonstrating the superiority of our method. Furthermore, we compare our method with other state-of-the-art (SOTA) methods. For instance, our NormAUG outperforms MVDG by +0.48\% (87.04 vs. 86.56) on ResNet-18. It is worth noting that MVDG generates multiple images through data augmentation for each test image and combines all the results for final prediction, which is also an ensemble scheme in the test stage.

\newcommand{\PreserveBackslash}[1]{\let\temp=\\#1\let\\=\temp}
\newcolumntype{C}[1]{>{\PreserveBackslash\centering}p{#1}}
\newcolumntype{R}[1]{>{\PreserveBackslash\raggedleft}p{#1}}
\newcolumntype{L}[1]{>{\PreserveBackslash\raggedright}p{#1}}
\begin{table}[htbp]
  \centering
\caption{Domain generalization accuracy~(\%) on PACS dataset with ResNet-18 (top) and ResNet-50 (bottom) backbone. The best performance is marked as \textbf{bold}, and the \underline{underline} is the second best result.}
    \begin{tabular}{l|cccc|c}
    \toprule
    \multicolumn{1}{c|}{Methods} & A     & C     & P     & S     & Avg. \\
    \midrule
   COMEN~\cite{DBLP:conf/cvpr/ChenL0LY22}  & ~82.60~ & ~81.00~ & ~94.60~ & ~84.50~ & 85.68 \\
    CIRL~\cite{DBLP:conf/cvpr/LvLLZLWL22}  & \textbf{86.08} & 80.59 & 95.93 & 82.67 & 86.32 \\
    I$^2$-ADR~\cite{DBLP:conf/eccv/MengLCYSWZSXP22}  & 82.90 & 80.80 & 95.00 & 83.50 & 85.55 \\
    XDED~\cite{DBLP:conf/eccv/LeeKK22}  & 85.60 & \textbf{84.20} & \underline{96.50} & 79.10 & 86.35 \\
    LRDG~\cite{ding2022domain} & 81.88 & 80.20 & 95.21 & 84.65 & 85.48 \\
     \midrule
    DAEL~\cite{zhou2021domain} &84.60 &74.40 &95.60 &78.90 &83.40 \\
    DSON~\cite{DBLP:conf/eccv/SeoSKKHH20}  & 84.67 & 77.65 & 95.87 & 82.23 & 85.11 \\
     BEN~\cite{DBLP:journals/pr/SeguTT23}  & 78.80 & 78.90 & 94.80 & 79.70 & 83.10 \\
      MVDG~\cite{DBLP:conf/eccv/Zhang0SG22}  & \underline{85.62} & 79.98 & 95.54 & \textbf{85.08} & \underline{86.56} \\
      \midrule
      MixStyle~\cite{DBLP:conf/iclr/ZhouY0X21} & 84.10 & 78.80 & 96.10 & 75.90 & 83.70\\
   EFDMix~\cite{DBLP:conf/cvpr/ZhangLLJZ22}  & 83.90 & 79.40 & \textbf{96.80} & 75.00 & 83.78 \\
      FACT~\cite{DBLP:conf/cvpr/XuZ0W021}  & 85.37 & 78.38 & 95.15 & 79.15 & 84.51 \\
     RSC~\cite{DBLP:conf/eccv/HuangWXH20}   & 83.43 & 80.31 & 95.99 & 80.85 & 85.15 \\
      STNP~\cite{DBLP:conf/cvpr/KangLKK22}  & 84.41 & 79.25 & 94.93 & 83.27 & 85.47 \\
    \midrule
    NormAUG (ours)& 85.60 & \underline{81.85} & 95.70 & \underline{85.00} & \textbf{87.04} \\
    \bottomrule
     \midrule
    LRDG~\cite{ding2022domain} & 86.57 & \underline{85.78} & 95.57 & \textbf{86.59} & 88.63 \\
    Fishr~\cite{DBLP:conf/icml/RameDC22} & 88.40 & 78.70 & 97.00 & 77.80 & 85.50 \\
    mDSDI~\cite{DBLP:conf/nips/BuiTTP21}  & 87.70 & 80.40 & \textbf{98.10} & 78.40 & 86.20 \\
    GIN~\cite{DBLP:journals/tip/XiaJD23} & 89.00 & 81.50 & \underline{98.00} & 80.20 & 87.20 \\
    CACE-D~\cite{DBLP:journals/tip/WangLCWHCH23} & 89.20 & 82.10 & \underline{98.00} & 80.50 & 87.50 \\
    I$^2$-ADR~\cite{DBLP:conf/eccv/MengLCYSWZSXP22} & 88.50 & 83.20 & 95.20 & 85.80 & 88.18 \\
    SWAD~\cite{DBLP:conf/nips/ChaCLCPLP21}  & 89.30 & 83.40 & 97.30 & 82.50 & 88.10 \\
     \midrule
    DSON~\cite{DBLP:conf/eccv/SeoSKKHH20}  & 87.04 & 80.62 & 95.99 & 82.90 & 86.64 \\
          MVDG~\cite{DBLP:conf/eccv/Zhang0SG22}  & 89.31 & 84.22 & 97.43 & \underline{86.36} & \underline{89.33} \\
           \midrule
    RSC~\cite{DBLP:conf/eccv/HuangWXH20}   & 87.89 & 82.16 & 97.92 & 83.35 & 87.83 \\
      FACT~\cite{DBLP:conf/cvpr/XuZ0W021}  & 89.63 & 81.77 & 96.75 & 84.46 & 88.15 \\
    STNP~\cite{DBLP:conf/cvpr/KangLKK22}  & \underline{90.35} & 84.20 & 96.73 & 85.18 & 89.12 \\
      EFDMix~\cite{DBLP:conf/cvpr/ZhangLLJZ22} & \textbf{90.60} & 82.50 & \textbf{98.10} & 76.40 & 86.90 \\
    \midrule
    NormAUG (ours) & 88.95 & \textbf{86.00} & 97.15 & 85.95 & \textbf{89.51} \\
    \bottomrule
    \end{tabular}%
  \label{tab01}%
\end{table}%

\textbf{Results on Office-Home.}
We also present the experimental results on the Office-Home dataset, as shown in Tab.~\ref{tab02}. As observed, our NormAUG method consistently outperforms the compared augmentation-based methods, including RSC~\cite{DBLP:conf/eccv/SeoSKKHH20}, MixStyle~\cite{DBLP:conf/iclr/ZhouY0X21}, L2A-OT~\cite{DBLP:conf/eccv/ZhouYHX20}, FACT~\cite{DBLP:conf/cvpr/XuZ0W021}, DSU~\cite{DBLP:conf/iclr/LiDGLSD22}, and STNP~\cite{DBLP:conf/cvpr/KangLKK22}. Additionally, we compare our method with the recent method DCG~\cite{lv2023improving}, further highlighting the effectiveness of our method.

\begin{table}[t]
 \centering
    \caption{Domain generalization accuracy~(\%) on Office-Home. The best performance is marked as \textbf{bold}, and the \underline{underline} is the second best result.}
      \begin{tabular}{l|cccc|c}
    \toprule
    \multicolumn{1}{c|}{Methods} & A     & C     & P     & R     & Avg. \\
    \midrule
     RSC~\cite{DBLP:conf/eccv/SeoSKKHH20}  & ~58.42~ & ~47.90~ & ~71.63~ & ~74.54~ & 63.12 \\
     MixStyle~\cite{DBLP:conf/iclr/ZhouY0X21}  & 58.70 & 53.40 & 74.20 & 75.90 & 65.55 \\
     L2A-OT~\cite{DBLP:conf/eccv/ZhouYHX20} & 60.60 & 50.10 & 74.80 & 77.00 & 65.63 \\
  FACT~\cite{DBLP:conf/cvpr/XuZ0W021}  & 60.34 & 54.85 & 74.48 & 76.55 & 66.56 \\
    DSU~\cite{DBLP:conf/iclr/LiDGLSD22}   & 60.20 & 54.80 & 74.10 & 75.10 & 66.05 \\
 STNP~\cite{DBLP:conf/cvpr/KangLKK22}  & 59.55 & 55.01 & 73.57 & 75.52 & 65.91 \\
      DAEL~\cite{zhou2021domain} & 59.40 & 55.10 & 74.00 & 75.70 & 66.10\\
    DCG~\cite{lv2023improving}   & \underline{60.67} & \underline{55.46} & \textbf{75.26} & \textbf{76.82} & \underline{67.05} \\
    \midrule
    NormAUG (ours) & \textbf{61.25} & \textbf{58.00} & \underline{75.25} &  \underline{76.65} & \textbf{67.79} \\
    \bottomrule
    \end{tabular}%
  \label{tab02}%
\end{table}

\textbf{Results on mini-DomainNet.} 
We further evaluate the effectiveness of our method on the mini-DomainNet dataset, which has a larger number of categories compared to PACS and Office-Home. In Tab.~\ref{tab03}, we report the experimental results and compare our method with various methods, including meta-learning, domain-invariant, augmentation, and other methods. As observed, our NormAUG method exhibits a clear advantage over all the compared methods, demonstrating its effectiveness on datasets with multiple classes.

\begin{table}[t]
 \centering
    \caption{Domain generalization accuracy~(\%) on  mini-DomainNet. The best performance is marked as \textbf{bold}, and the \underline{underline} is the second best result.}
    \begin{tabular}{l|cccc|c}
    \toprule
    \multicolumn{1}{c|}{Methods} & C     & P     & R     & S     & Avg. \\
    \midrule
    MLDG~\cite{DBLP:conf/aaai/LiYSH18}  & ~65.70~ & ~57.00~ & ~63.70~ & ~58.10~ & 61.12 \\
    DAEL~\cite{zhou2021domain} & 69.95 & 55.13 & 66.11& 55.72& 61.73\\
    MMD~\cite{DBLP:conf/cvpr/LeeBBU19}   & 65.00 & 58.00 & 63.80 & 58.40 & 61.30 \\
    Mixup~\cite{DBLP:conf/iclr/ZhangCDL18} & 67.10 & 59.10 & 64.30 & 59.20 & 62.43 \\
    SagNet~\cite{DBLP:conf/cvpr/NamLPYY21} & 65.00 & 58.10 & 64.20 & 58.10 & 61.35 \\
    CORAL~\cite{DBLP:conf/iccv/PengBXHSW19} & 66.50 & 59.50 & 66.00 & 59.50 & 62.87 \\
    MTL~\cite{DBLP:journals/jmlr/BlanchardDDLS21}   & 65.30 & 59.00 & 65.60 & 58.50 & 62.10 \\
    DCG~\cite{lv2023improving}   & \underline{69.38} & \underline{61.79} & \underline{66.34} & \underline{63.21} & \underline{65.18} \\
    \midrule
    NormAUG (ours) & \textbf{70.20} & \textbf{66.90} & \textbf{71.20} & \textbf{63.40} & \textbf{67.93} \\
    \bottomrule
     \end{tabular}%
  \label{tab03}%
\end{table}

\textbf{Results on Digits-DG.} 
We also evaluate the performance of our method on the Digits-DG dataset, which is a widely used benchmark dataset for domain generalization in digit recognition. The experimental results are presented in Tab.~\ref{tab04}.
As observed in the table, our method consistently outperforms the other methods, highlighting its effectiveness in the domain generalization task for digit recognition. The superior performance of our method on this dataset further validates its robustness and generalization capability.

\begin{table}[htbp]
  \centering
   \caption{Domain generalization accuracy~(\%) on Digits-DG. The best performance is marked as \textbf{bold}, and the \underline{underline} is the second best result.}
    \begin{tabular}{l|cccc|c}
    \toprule
    \multicolumn{1}{c|}{Methods} & MN    & MN-M  & SV    & SY    & Avg. \\
    \midrule
    FACT~\cite{DBLP:conf/cvpr/XuZ0W021}  & ~\textbf{97.90}~ & ~65.60~ & ~72.40~ & ~90.30~ & 81.55 \\
    COMEN~\cite{DBLP:conf/cvpr/ChenL0LY22}  & 97.10 & 67.60 & 75.10 & 91.30 & 82.78 \\
    CIRL~\cite{DBLP:conf/cvpr/LvLLZLWL22}  & 96.08 & \textbf{69.87} & \underline{76.17} & 87.68 & 82.45 \\
    STEAM~\cite{DBLP:conf/iccv/Chen0PY0021}  & 96.80 & 67.50 & 76.00 & \underline{92.20} & \underline{83.13} \\
    \midrule
    NormAUG (ours) & \underline{97.50} & \underline{67.65} & \textbf{80.10} & \textbf{96.85} & \textbf{85.53} \\
    \bottomrule
    \end{tabular}%
  \label{tab04}%
\end{table}%

\textbf{Results on DomainNet.}
We evaluate the performance of our method on the DomainNet dataset, which has more domains compared to the other datasets in our experiment. The results are reported in Tab.~\ref{tab11}. We can observe that on large dataset with multiple source domains, our method exhibits a remarkably substantial improvement in performance.

\newcolumntype{C}[1]{>{\PreserveBackslash\centering}p{#1}}
\newcolumntype{R}[1]{>{\PreserveBackslash\raggedleft}p{#1}}
\newcolumntype{L}[1]{>{\PreserveBackslash\raggedright}p{#1}}
\begin{table}[t]
 \centering
    \caption{Domain generalization accuracy~(\%) on DomainNet with ResNet-18 (top) and ResNet-50 (bottom) backbone. The best performance is marked as \textbf{bold}, and the \underline{underline} is the second best result.}
      \begin{tabular}{L{2.01cm}|C{0.45cm}C{0.45cm}C{0.45cm}C{0.45cm}C{0.45cm}C{0.45cm}|C{0.45cm}}
    \toprule
    \multicolumn{1}{c|}{Methods} & C     & I     & P     & Q     &R     &S     & Avg. \\
    \midrule
     MetaReg ~\cite{DBLP:conf/nips/BalajiSC18}  & 53.70 & \underline{21.10}  & \underline{45.30}  & 10.60 & \textbf{58.50} & 42.30 & 38.58 \\
     DMG ~\cite{DBLP:conf/eccv/ChattopadhyayBH20} & \textbf{60.10} & 18.80  & 44.50  & \underline{14.20} & 54.70 & 41.70 & \underline{39.00} \\
     I$^2$-ADR~\cite{DBLP:conf/eccv/MengLCYSWZSXP22} & 57.30 & 15.20  & 44.10  & 12.10 & 53.90 & \underline{46.70} & 38.22 \\
     ITTA~\cite{Chen_2023_CVPR} &  50.70 & 13.90  &  39.40  &  11.90 &  50.20 & 43.50 & 34.90 \\
    \midrule
    
    NormAUG (ours) & \underline{57.40} & \textbf{22.70} & \textbf{49.00} & \textbf{14.60} & \underline{58.30} &  \textbf{48.70} & \textbf{41.78} \\
    \bottomrule
    \midrule
     RSC~\cite{DBLP:conf/eccv/HuangWXH20} & 55.00 & 18.30  & 44.40  & 12.20 & 55.70 & 47.80 & 38.90 \\
     DMG ~\cite{DBLP:conf/eccv/ChattopadhyayBH20} & \textbf{65.20} & \underline{22.20}  & 50.00  & \underline{15.70} & 59.60 & 49.00 & 43.62 \\
     SagNet~\cite{DBLP:conf/cvpr/NamLPYY21} & 57.70 & 19.00  & 45.30  & 12.70 & 58.10 & 48.80 & 40.27 \\
     SelfReg ~\cite{DBLP:conf/iccv/KimYPKL21} & 60.70 & 21.60  & 49.40  & 12.70 & 60.70  &51.70  & 42.80 \\
     I$^2$-ADR~\cite{DBLP:conf/eccv/MengLCYSWZSXP22} & 64.40 & 20.20  & 49.20  & 15.00 & 61.60 & 53.30 & 43.95 \\
     PTE ~\cite{DBLP:conf/eccv/MinPKPK22} & 62.40 & 21.00  &  50.50  &  13.80 &  \textbf{64.60} & 52.40 & 44.12 \\
     SAGM ~\cite{Wang_2023_CVPR} & \underline{64.90} & 21.10 & \underline{51.50} & 14.80 & \underline{64.10} & \underline{53.60} & \underline{45.00} \\
    \midrule
    
    NormAUG (ours) & 63.10 & \textbf{27.30} & \textbf{54.30} & \textbf{17.30} & 62.00 &  \textbf{54.80} & \textbf{46.47} \\
    \bottomrule
    \end{tabular}%
  \label{tab11}%
\end{table}

\subsection{Ablation Study}~\label{sec:EXP-SS}
 We conducted an ablation study on the PACS and mini-DomainNet datasets to analyze the effectiveness of different modules. The experimental results are presented in Tab.~\ref{tab05}. In this table, ``ON'' represents the optimized normalization~\cite{DBLP:conf/eccv/SeoSKKHH20} module in the main path, ``AUG'' represents the normalization-based augmentation module in the training stage, and ``EP'' represents the ensemble prediction based on the normalization-based augmentation module.

As observed in the table, the ``ON'' module is effective for the domain generalization task as it incorporates the instance normalization scheme within the normalization layer. It contributes to improved performance. Furthermore, the addition of the ``AUG'' module enhances the diversity of samples and further boosts the performance. Finally, the combination of the ensemble prediction (``EP'') in the test stage leads to additional performance improvement.

Overall, the results demonstrate the effectiveness of each module and the benefits of combining them for improved domain generalization performance.

  \newcolumntype{C}[1]{>{\PreserveBackslash\centering}p{#1}}
\newcolumntype{R}[1]{>{\PreserveBackslash\raggedleft}p{#1}}
\newcolumntype{L}[1]{>{\PreserveBackslash\raggedright}p{#1}}

\begin{table}[htbp]
  \centering
  \caption{Ablation Studies on PACS and mini-DomainNet. In this table, ``DA'' denotes the ``DeepAll'' model.}
    \begin{tabular}{l|C{0.2cm}C{0.3cm}C{0.3cm}|cccc|c}
    \toprule
    \multicolumn{1}{c|}{\multirow{2}[1]{*}{Methods}} & \multirow{2}[1]{*}{ON} & \multirow{2}[1]{*}{AUG} & \multirow{2}[1]{*}{EP} & \multicolumn{5}{c}{PACS} \\
\cmidrule{5-9}          &       &       &       & A     & C     & P     & S     & Avg. \\
    \midrule
    DA~\cite{lv2023improving} &   -    &     -     &    -      & 77.63 & 76.77 & 95.85 & 69.50 & 79.94 \\
    \midrule
    Model-1 &    $\checkmark$     &   -       &     -     & 80.64 & 81.68 & 95.30 & 79.34 & 84.24 \\
    Model-2 &    $\checkmark$     &    $\checkmark$     &    -      & 83.85 & 82.55 & 94.95 & 83.85 & 86.30 \\
       \rowcolor[rgb]{ .851,  .851,  .851}  Ours &     $\checkmark$    &  $\checkmark$       &    $\checkmark$     & 85.60 & 81.85 & 95.70 & 85.00 & 87.04 \\
    \midrule
    \multicolumn{1}{c|}{\multirow{2}[1]{*}{Method}} & \multirow{2}[1]{*}{ON} & \multirow{2}[1]{*}{AUG} & \multirow{2}[1]{*}{EP} & \multicolumn{5}{c}{mini-DomainNet} \\
\cmidrule{5-9}          &       &       &       & C     & P     & R     & S     & Avg. \\
    \midrule
    DA~\cite{lv2023improving} &     -     &     -     &    -      & 65.30 & 58.40 & 64.70 & 59.00 & 61.86 \\
    \midrule
    Model-1 &     $\checkmark$    &    -      &    -      & 67.95 & 63.10 & 70.00 & 57.10 & 64.54 \\
    Model-2 &    $\checkmark$     &   $\checkmark$      &     -     & 69.80 & 65.80 & 69.90 & 61.70 & 66.80 \\
       \rowcolor[rgb]{ .851,  .851,  .851}  Ours &    $\checkmark$     &    $\checkmark$     &     $\checkmark$    & 70.20 & 66.90 & 71.20 & 63.40 & 67.93 \\
    \bottomrule
    \end{tabular}%
  \label{tab05}%
  \vspace{-15pt}
\end{table}%

\subsection{Further Analysis}~\label{sec:EXP-FA}
\begin{figure*}
\centering
\subfigure[PACS]{
\includegraphics[width=4.0cm]{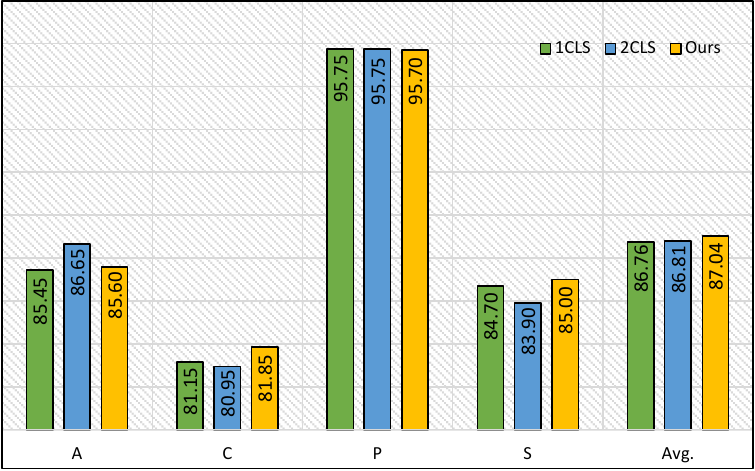}
}
\subfigure[mini-DomainNet]{
\includegraphics[width=4.0cm]{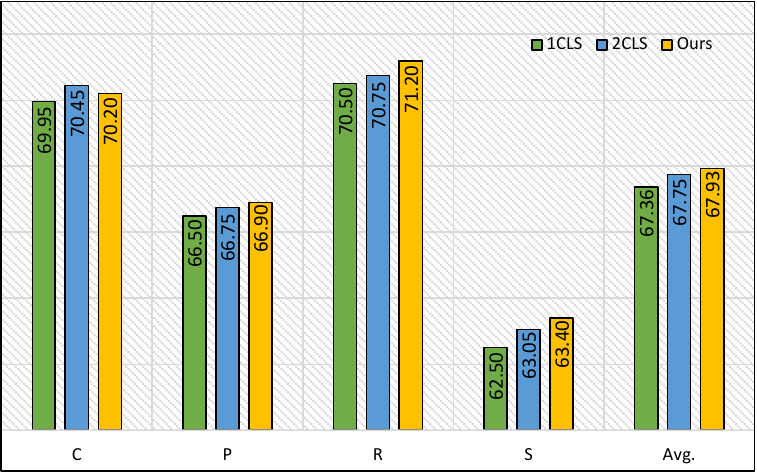}
}
\subfigure[Office-Home]{
\includegraphics[width=4.0cm]{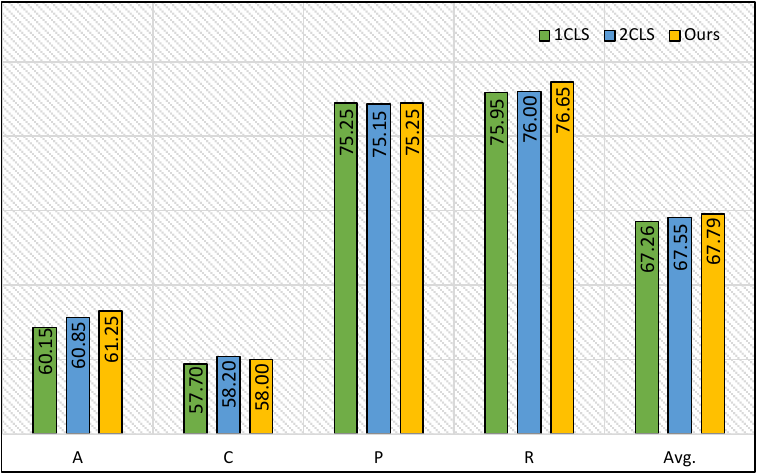}
}
\subfigure[Digit\_DG]{
\includegraphics[width=4.0cm]{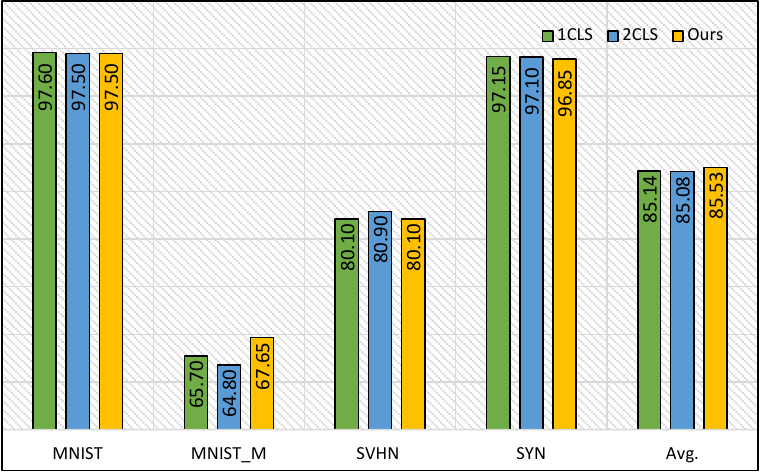}
}
\caption{Experimental results of different numbers of classifiers in our method on four datasets. In this figure, ``1CLS'' denotes that the main path and the auxiliary path share the same classifier. ``2CLS'' indicates that the main path and the auxiliary path use different classifiers, and these sub-paths in the auxiliary path share a classifiers. It is worth noting that all method use the same fused scheme in the test stage.}
\label{fig01}
  \vspace{-15pt}
\end{figure*}

\textbf{Impact on different numbers of classifiers.}
As mentioned in Section~\ref{s-framework}, we explore different configurations for the classifiers in the training process. Specifically, we consider two cases: one shared classifier and two classifiers. In the case of two classifiers, one is assigned to the main path, while the other is assigned to the auxiliary path (\ie, all sub-paths in the auxiliary path share a classifier). It is important to note that the experimental setup and the fused scheme used in the test stage are consistent across all variant methods. The experimental results are summarized in Fig.~\ref{fig01}.
As observed in the figure, using independent classifiers for each normalization combination yields the best performance among the variant methods. This can be attributed to the increased diversity introduced by the multiple different classifiers. Therefore, our method adopts the use of independent classifiers for all paths. It is important to emphasize that all methods in Fig.~\ref{fig01} are evaluated using the same fused scheme in the test stage.
Furthermore, we investigate the impact of the number of classifiers for the main path, as reported in Tab.~\ref{tab10}. On the PACS dataset, using one classifier (``1CLS'') and two classifiers (``2CLS'') for the main path results in a slight performance decrease of $-0.22\%$ and $-0.26\%$, respectively, compared to our method. Similarly, on the mini-DomainNet dataset, ``1CLS'' and ``2CLS'' lead to a performance decrease of $-0.41\%$ and $-0.00\%$ compared to our method in the main path. These results indicate that employing different classifiers for each BN in the BN bank can also slightly improve the performance of the main path.

\begin{table}[htbp]
  \centering
  \caption{Experimental results of the main path. The model uses different numbers of classifiers, which is corresponding with Fig~\ref{fig01} (a) and (b). ``Drop'' in this table denotes the value of ``1CLS or 2CLS-Ours''.}
    \begin{tabular}{l|cccc|c|c}
    \toprule
    \multicolumn{7}{c}{PACS} \\
    \midrule
    \multicolumn{1}{c|}{~~Methods~~} & A     & C     & P     & S     & ~~Avg.~~  & ~Drop~ \\
    \midrule
    \rowcolor[rgb]{ .851,  .851,  .851} Ours  & 83.85 & 82.55 & 94.95 & 83.85 & 86.30 & - \\
    1CLS  & 83.25 & 82.15 & 95.05 & 83.85 & 86.08 &  -0.22\\
    2CLS  & 84.20 & 81.85 & 95.00 & 83.10 & 86.04 &  -0.26\\
    \midrule
    \multicolumn{7}{c}{mini-DomainNet} \\
    \midrule
    \multicolumn{1}{c|}{Methods} & C     & P     & R     & S     & Avg.  & Drop \\
    \midrule
    \rowcolor[rgb]{ .851,  .851,  .851} Ours  & 69.80 & 65.80 & 69.90 & 61.70 & 66.80 & - \\
    1CLS  & 69.50 & 65.45 & 69.35 & 61.25 & 66.39 & -0.41 \\
    2CLS  & 70.05 & 65.70 & 69.55 & 61.90 & 66.80 &  -0.00\\
    \bottomrule
    \end{tabular}%
  \label{tab10}
  \vspace{-5pt}
\end{table}%

\textbf{Evaluation on different fusion schemes on prediction.}
In the test phase, we combine the results from the main path and the independent domain paths. In our method, recognizing the importance of the main path, we adopt a two-step averaging method. First, we average the results from the independent domain paths, and then we average this averaged result with the result from the main path. Additionally, we also evaluate a direct averaging method where we average all results from the main path and the independent domain paths. The experimental results are presented in Tab.~\ref{tab06}.
As observed in the table, using the two-step averaging scheme leads to a decrease in performance compared to the direct averaging method. This suggests that the two-step averaging scheme reduces the influence of the main path's result in the final ensemble result.
\begin{table}[htbp]
  \centering
  \caption{Experimental results of different fusion schemes on prediction.}
    \begin{tabular}{l|cccc}
    \toprule
    \multicolumn{1}{c|}{Methods} & ~PACS~  & Office-Home & mini-DomainNet & Digit\_DG \\
    \midrule
    MeanAll & 86.21 & 67.70 & 67.88 & 85.33 \\
        \rowcolor[rgb]{ .851,  .851,  .851} Ours  & 87.04 & 67.79 & 67.93 & 85.53 \\
    \bottomrule
    \end{tabular}%
  \label{tab06}%
  \vspace{-5pt}
\end{table}%

\textbf{Analysis of the divergence.} 
In this section, we evaluate the effectiveness of our NormAUG method in reducing the domain gap between source and source ($\mathcal{D}{s2s}$) and between source and target ($\mathcal{D}{s2t}$). To compute these divergences, we calculate the averaged feature of the $d$-th domain, denoted as $\bar{f^s_d}$, the averaged feature of the source domains, denoted as $\bar{f^s}$, and the averaged feature of the target domain, denoted as $\bar{f^t}$.
By utilizing these features, we can compute the divergence between source and source as $\mathcal{D}_{s2s}=\frac{1}{D}\sum_{d=1}^{D}\|\bar{f^s}-\bar{f^s_d}\|$ and the divergence between source and target as $\mathcal{D}_{s2t}=\|\bar{f^s}-\bar{f^t}\|$, where $D$ is the number of source domains.
The experimental results are presented in Tab.~\ref{tab07}. In this table, ``Model-1'' is the same as ``Model-1'' in Tab.~\ref{tab05}. As observed, our NormAUG method effectively reduces the domain gap between source and source as well as between source and target. This demonstrates that our method is capable of learning domain-invariant features, thus enhancing the generalization capability across different domains.

\begin{table}[htbp]
  \centering
  \caption{Data-distribution distance between source and source (S2S) and between source and target (S2T) on PACS. Here, a smaller domain gap is preferable.}
    \begin{tabular}{c|cc|cc}
    \toprule
    \multirow{2}[1]{*}{Tasks} & \multicolumn{2}{c|}{S2T} & \multicolumn{2}{c}{S2S} \\
\cmidrule{2-5}          & ~~Ours~~  & ~~Model-1~~ & ~~Ours~~  & ~~Model-1~~ \\
    \midrule
    ~~CPS$\to$A~~ & 1.82  & 2.49  & 1.83  & 1.91 \\
    APS$\to$C & 1.88  & 2.40  & 1.79  & 1.88 \\
    ACS$\to$P & 3.49  & 3.76  & 1.59  & 1.84 \\
    ACP$\to$S & 4.62  & 4.79  & 1.65  & 1.67 \\
    \bottomrule
    \end{tabular}%
  \label{tab07}%
  \vspace{-0pt}
\end{table}%

\textbf{Evaluation on ``AUG-1'' in the BN bank.}
As discussed in Section~\ref{s-framework}, our NormAUG method incorporates a random combination scheme for conducting the normalization operation in the auxiliary at each iteration. To evaluate the effectiveness of this scheme, we always perform experiments using a single domain to independently conduct the normalization, which is the same as the test scheme in Fig.~\ref{fig06} or ``AUG-1'' in Fig.~\ref{fig05}. The experimental results are presented in Tab.~\ref{tab08}.
From the results, we can observe that the random combination scheme significantly improves the model's generalization compared to using a single domain for normalization. This improvement can be attributed to the increased diversity introduced by the random combination scheme. The random combination allows the model to explore different combinations of multiple domains during the normalization process, leading to a more comprehensive representation of the data and enhancing the model's ability to generalize across domains.

\begin{table}[htbp]
  \centering
  \caption{Comparison between the single-domain augmentation and random combination augmentation.}
    \begin{tabular}{l|cccc|c}
    \toprule
    \multicolumn{1}{c|}{\multirow{2}[1]{*}{~Methods~}} & \multicolumn{5}{c}{PACS} \\
\cmidrule{2-6}          & A     & C     & P     & S     & Avg. \\
    \midrule
    Single & ~85.28~ & ~79.54~ & ~96.66~ & ~83.10~ & ~86.15~ \\
          \rowcolor[rgb]{ .851,  .851,  .851}   Ours  & 85.60 & 81.85 & 95.70 & 85.00 & 87.04 \\
    \midrule
    \multicolumn{1}{c|}{\multirow{2}[1]{*}{Methods}} & \multicolumn{5}{c}{Office-Home} \\
\cmidrule{2-6}    
    & A     & C     & P     & R     & Avg. \\
    \midrule   
    Single & 60.60 & 56.40 & 73.60 & 76.20 & 66.70 \\
        \rowcolor[rgb]{ .851,  .851,  .851} Ours  & 61.25 & 58.00 & 75.25 & 76.65 & 67.79 \\
    \midrule
    \multicolumn{1}{c|}{\multirow{2}[1]{*}{Methods}} & \multicolumn{5}{c}{mini-DomainNet} \\
\cmidrule{2-6}          & C     & P     & R     & S     & Avg. \\
    \midrule
    Single & 68.40 & 65.90 & 70.90 & 61.90 & 66.78 \\
        \rowcolor[rgb]{ .851,  .851,  .851} Ours  & 70.20 & 66.90 & 71.20 & 63.40 & 67.93 \\
    \midrule
    \multicolumn{1}{c|}{\multirow{2}[1]{*}{Methods}} & \multicolumn{5}{c}{Digit\_DG} \\
\cmidrule{2-6}          & MN    & MN-M  & SV    & SY    & Avg. \\
    \midrule
    Single & 97.45 & 67.10 & 79.68 & 96.40 & 85.16 \\
        \rowcolor[rgb]{ .851,  .851,  .851} Ours  & 97.50 & 67.65 & 80.10 & 96.85 & 85.53 \\
    \bottomrule
    \end{tabular}%
  \label{tab08}%
   \vspace{-0pt}
\end{table}%
\textbf{Result of using all sub-paths in the test stage.} 
We also conduct experiments using all sub-paths of the auxiliary path in the test stage. The experimental results are presented in Tab.~\ref{tab09}. From this table, we observe that using all sub-paths of the auxiliary path (\ie, ``All'' in Tab.~\ref{tab09}) can slightly improve the performance compared to using the sub-paths for the independent domains alone, especially on the PACS and Office-Home datasets. However, it is important to note that using all sub-paths requires additional computation cost in the test stage.
In our NormAUG method, we primarily focus on utilizing the sub-paths for the independent domain, which has shown significant improvements in performance, as demonstrated in Tab.~\ref{tab05}. This method strikes a balance between performance improvement and computational efficiency. By utilizing the sub-paths for the independent domain, we can effectively enhance the model's generalization while maintaining a reasonable computational cost in the test stage.

\begin{table}[htbp]
  \centering
  \caption{Experimental results with different sub-paths of the auxiliary path in the test stage. ``All'' is using all sub-paths, and ``Independent'' is only using these sub-paths for the independent domain in the test stage.}
    \begin{tabular}{c|ccccc}
    \toprule
    \multicolumn{6}{c}{PACS} \\
    \midrule
    Methods & A     & C     & P     & S     & Avg. \\
    \midrule
    All   & 86.05 & 82.25 & 95.80 & 84.80 & 87.23 \\
    \rowcolor[rgb]{ .851,  .851,  .851} Independent (ours) & 85.60 & 81.85 & 95.70 & 85.00 & 87.04 \\
    \midrule
    \multicolumn{6}{c}{Office-Home} \\
    \midrule
    Methods & A     & C     & P     & R     & Avg. \\
    \midrule
    All   & 61.20 & 58.15 & 75.30 & 76.75 & 67.85 \\
    \rowcolor[rgb]{ .851,  .851,  .851} Independent (ours) & 61.25 & 58.00 & 75.25 & 76.65 & 67.79 \\
    \midrule
    \multicolumn{6}{c}{mini-DomainNet} \\
    \midrule
    Methods & C     & P     & R     & S     & Avg. \\
    \midrule
    All   & 70.40 & 66.80 & 71.20 & 63.20 & 67.90 \\
    \rowcolor[rgb]{ .851,  .851,  .851} Independent  (ours) & 70.20 & 66.90 & 71.20 & 63.40 & 67.93 \\
    \bottomrule
    \end{tabular}%
  \label{tab09}%
\end{table}%
\begin{figure*}
\centering
\subfigure[P+C+S$\to$A (PACS)]{
\includegraphics[width=4.12cm]{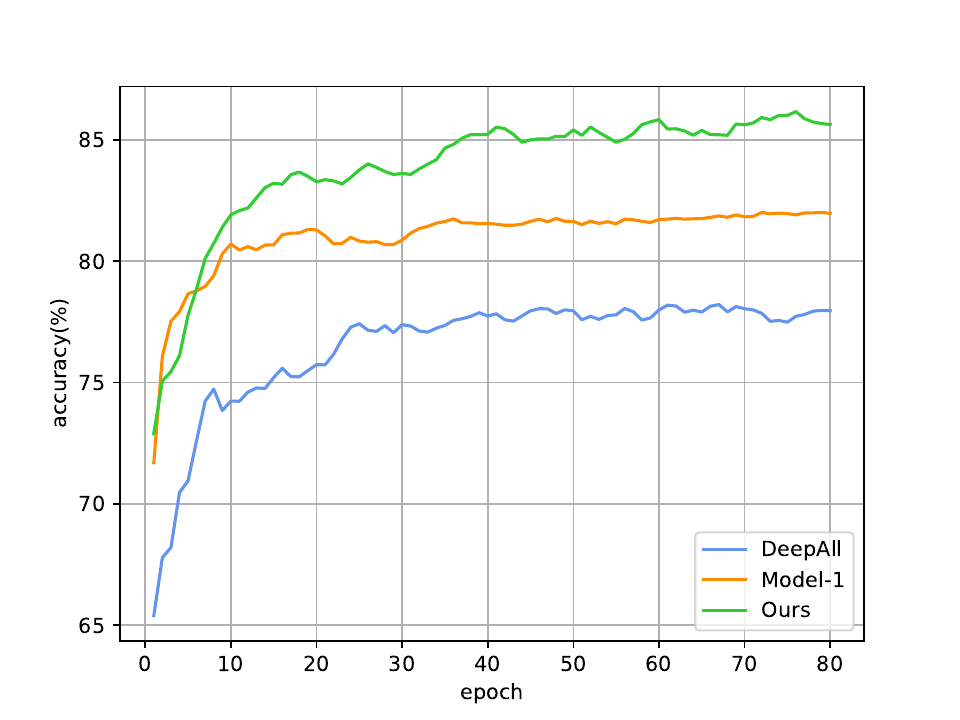}
}
\subfigure[P+A+S$\to$C (PACS)]{
\includegraphics[width=4.12cm]{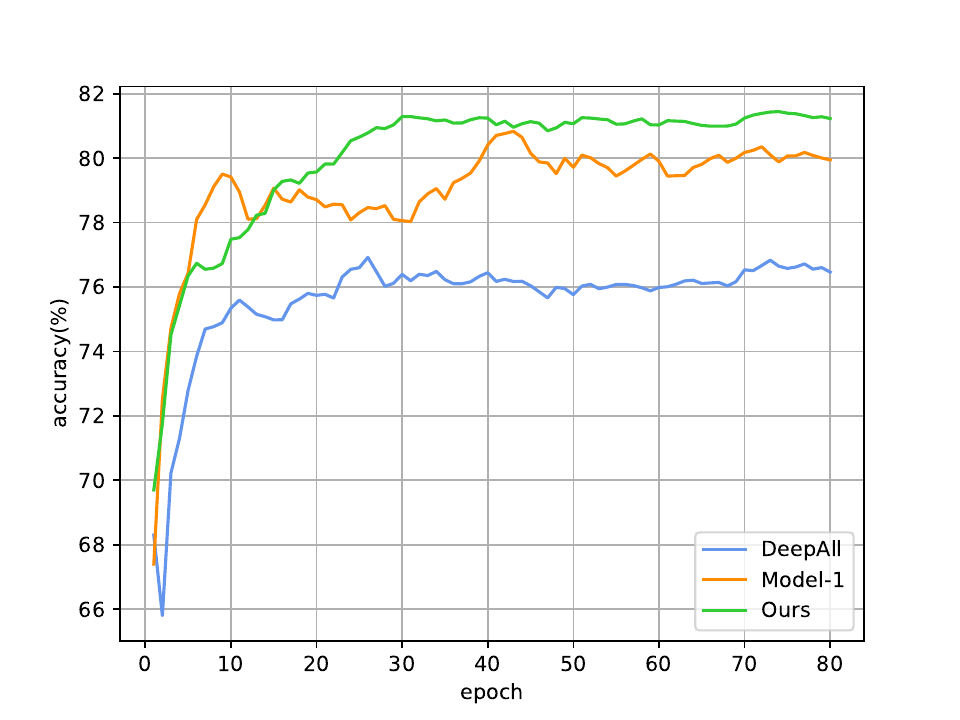}
}
\subfigure[A+C+S$\to$P (PACS)]{
\includegraphics[width=4.12cm]{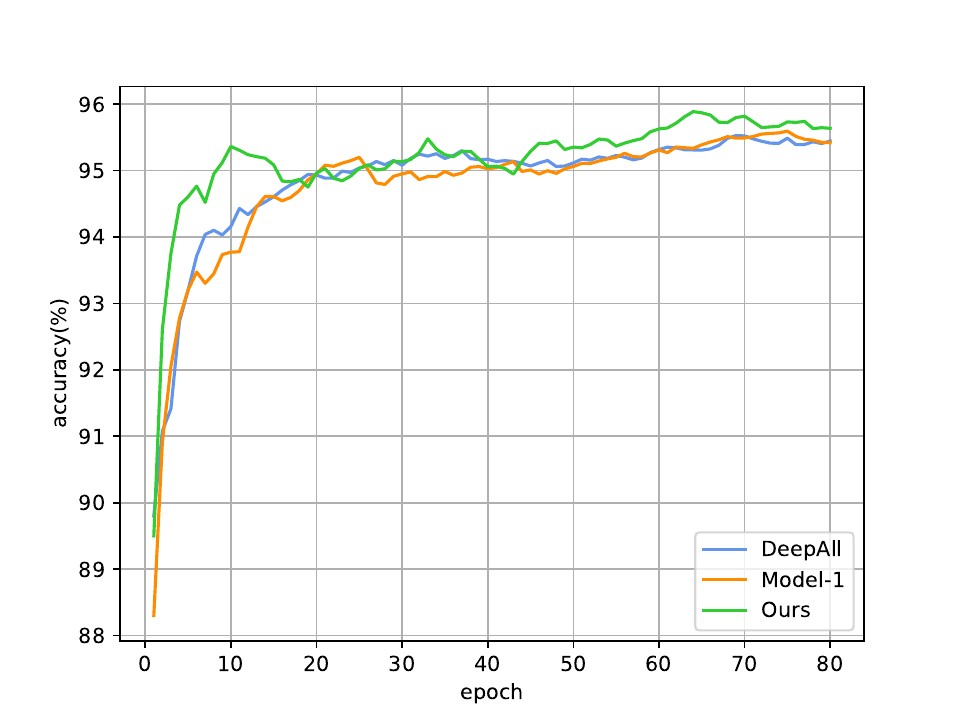}
}
\subfigure[P+C+A$\to$S (PACS)]{
\includegraphics[width=4.12cm]{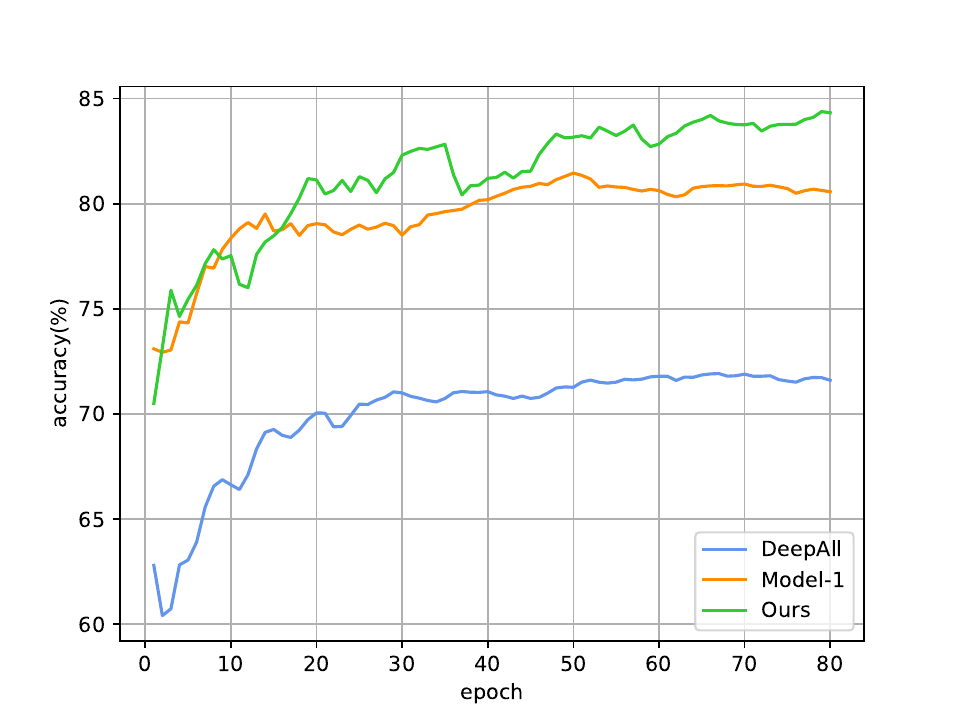}
}
\subfigure[P+C+R$\to$A (OH)]{
\includegraphics[width=4.12cm]{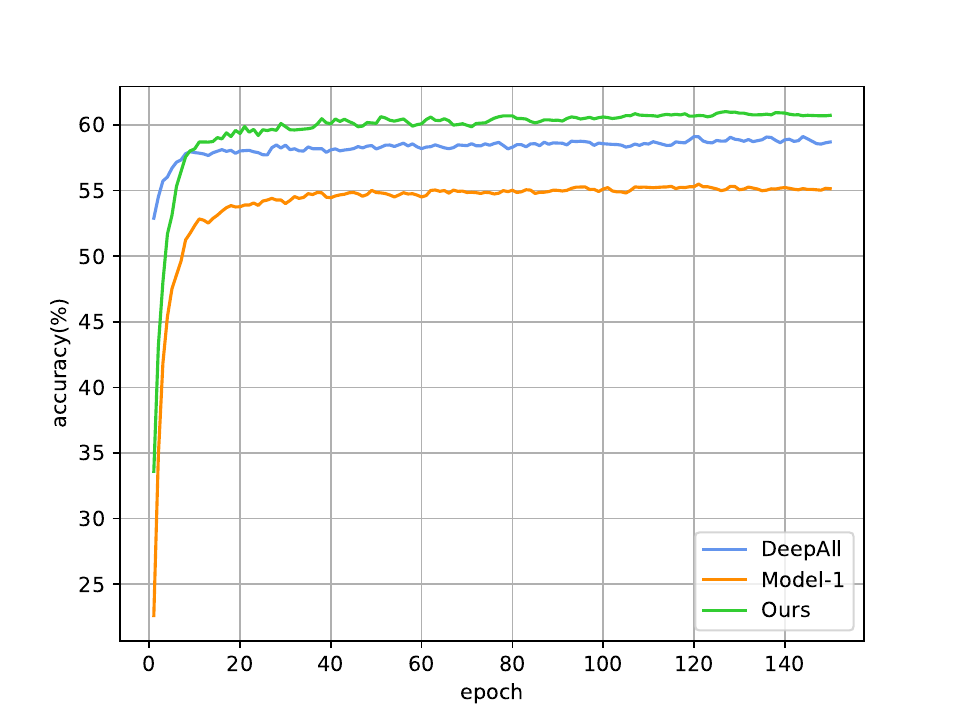}
}
\subfigure[P+A+R$\to$C (OH)]{
\includegraphics[width=4.12cm]{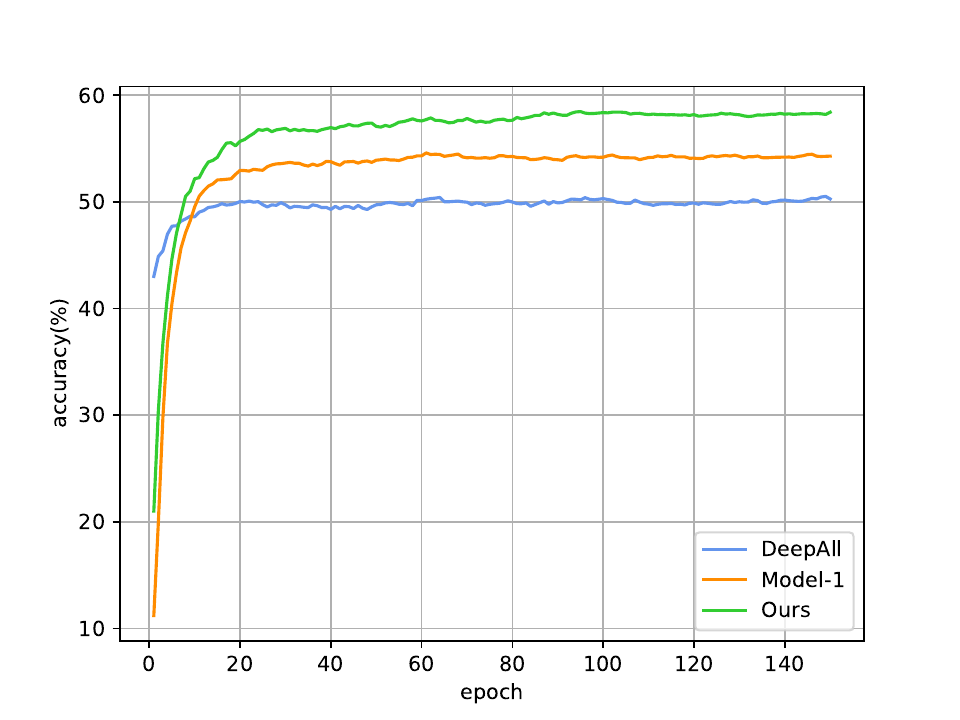}
}
\subfigure[A+C+R$\to$P (OH)]{
\includegraphics[width=4.12cm]{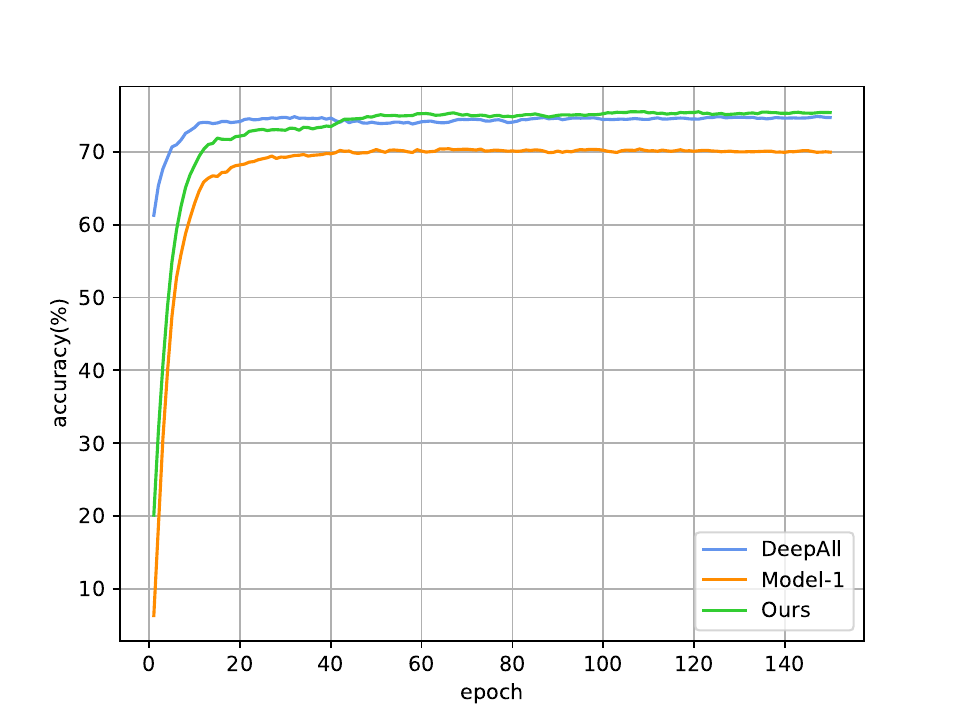}
}
\subfigure[P+C+A$\to$R (OH)]{
\includegraphics[width=4.12cm]{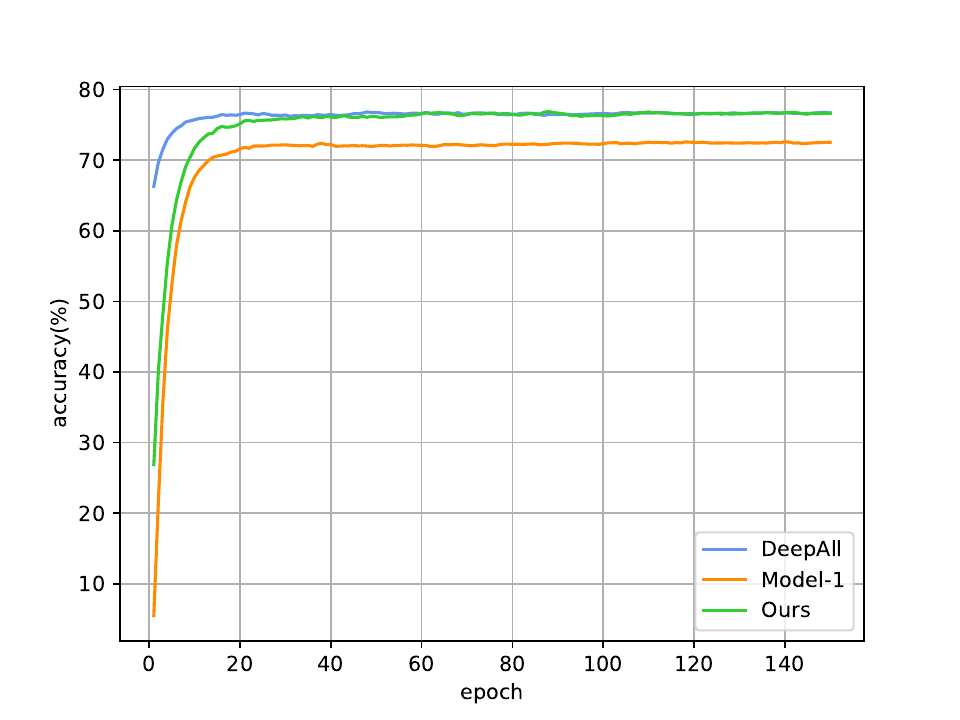}
}
\caption{The stability of the training process on PACS and Office-Home (OH).}
\label{fig02}
\vspace{-0pt}
\end{figure*}
\begin{figure*}
\centering
\subfigure[Sketch]{
\includegraphics[width=8.5cm]{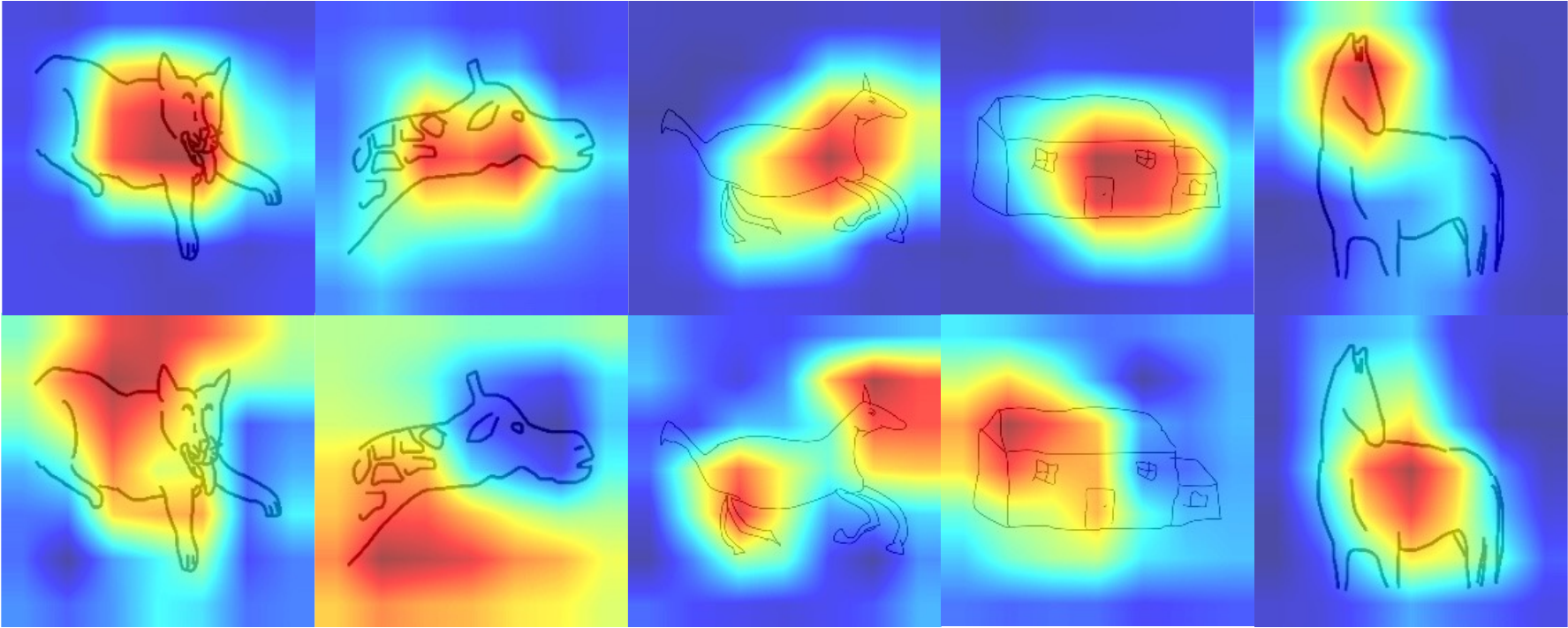}
}
\subfigure[Photo]{
\includegraphics[width=8.5cm]{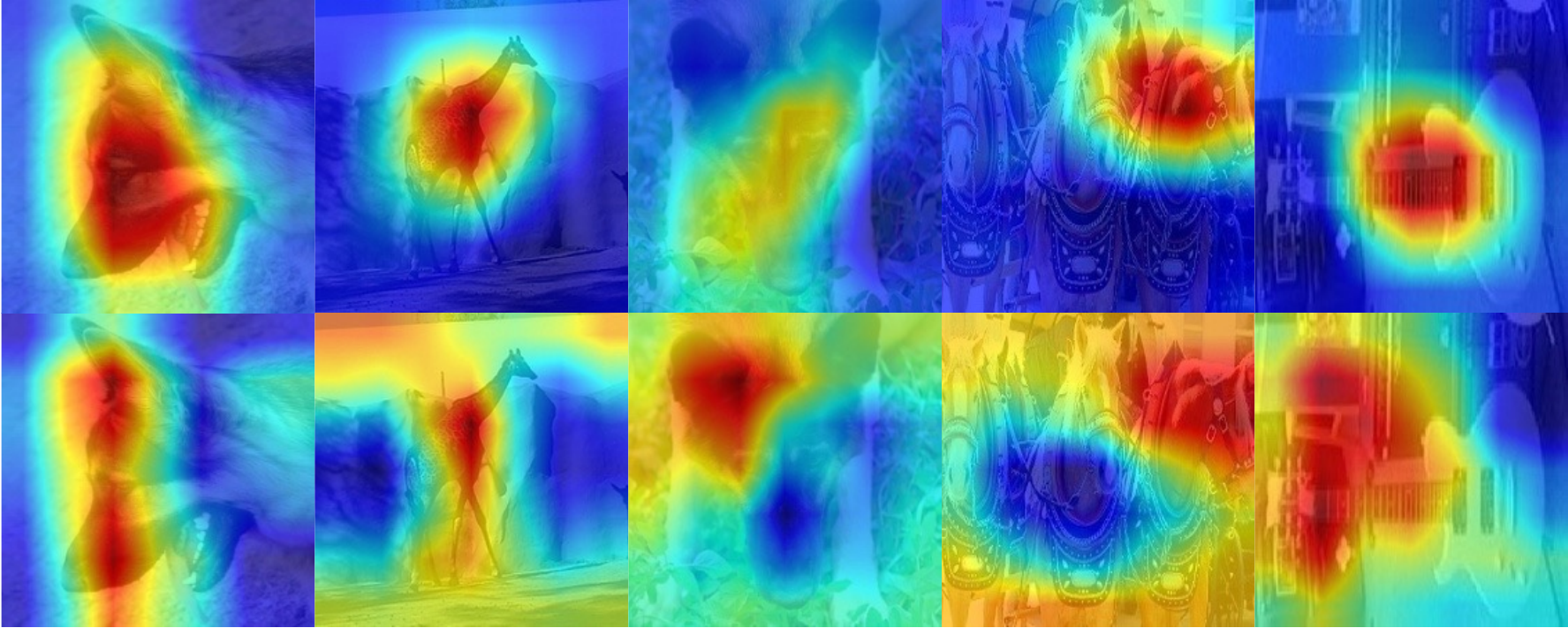}
}
\subfigure[Cartoon]{
\includegraphics[width=8.5cm]{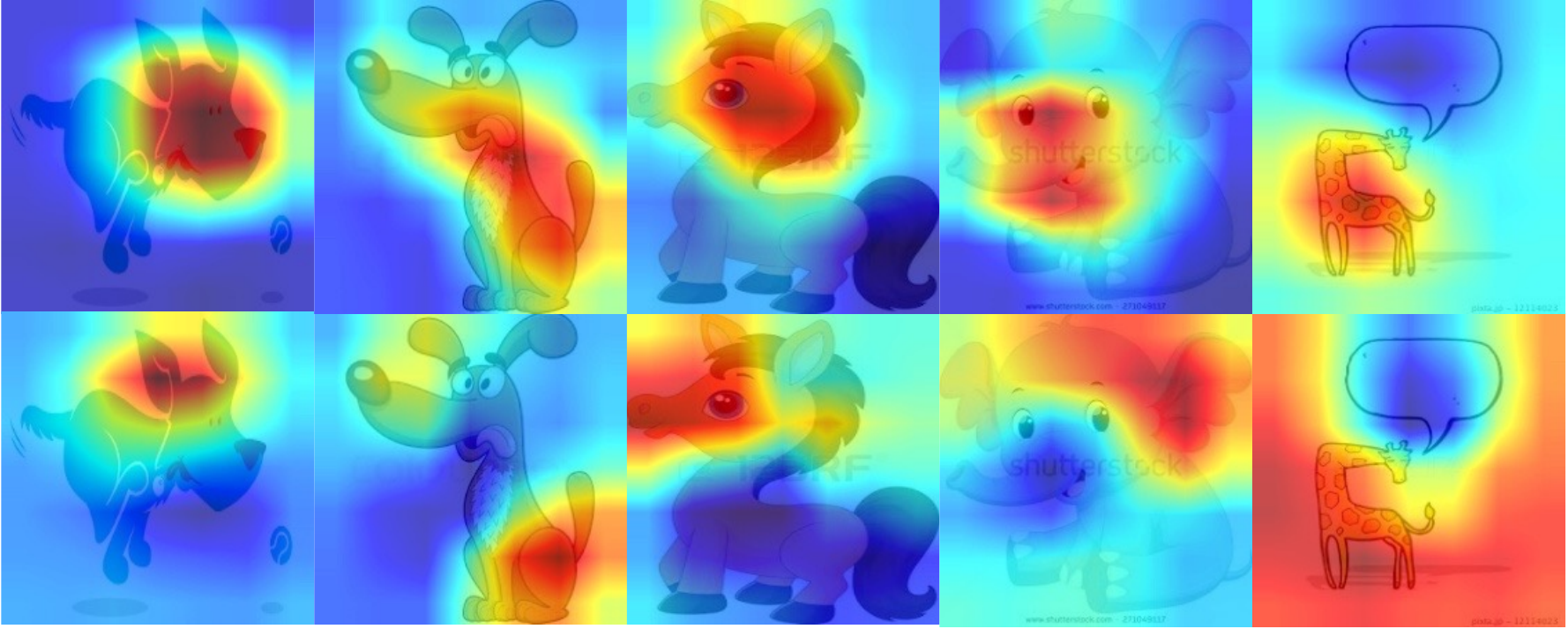}
}
\subfigure[Art painting]{
\includegraphics[width=8.5cm]{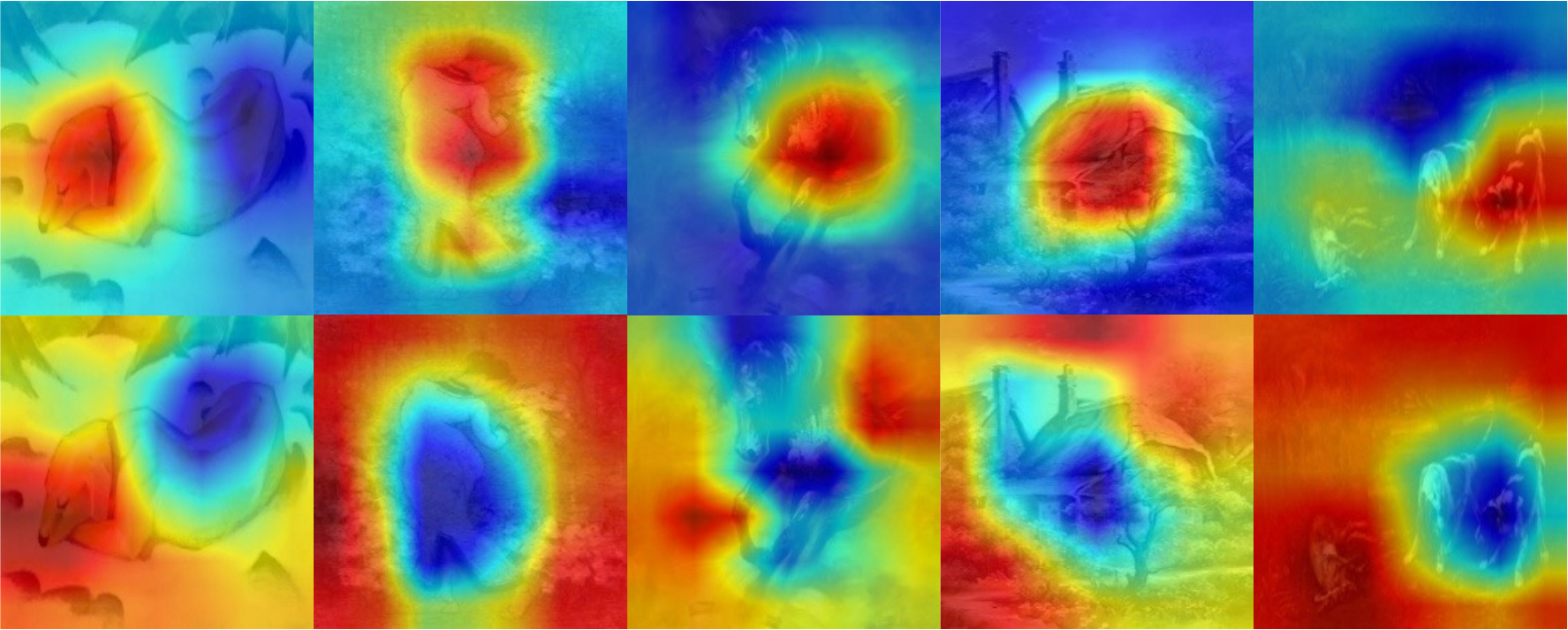}
}
\caption{The activation maps of our method (top) and the baseline (bottom) on PACS. In this figure, the redder area indicates the more attention.
}
\label{fig03}
\vspace*{-10pt}
\end{figure*}
\textbf{Stability of the training process.} 
We demonstrate the stability of our method during the training process, as illustrated in Fig.~\ref{fig02}. From these figures, it is evident that our method exhibits a consistent improvement in performance over time, specifically in the unseen target domain. This stability further emphasizes the effectiveness and robustness of our method in enhancing the model's generalization capabilities. Different from multiple  existing methods that select the best model for evaluation~\cite{DBLP:conf/eccv/MengLCYSWZSXP22,DBLP:conf/eccv/Zhang0SG22}, we leverage the model from the last one epoch for evaluation in all experiments.

\textbf{Evaluation of the test time.}
Our method introduces a Batch Normalization (BN) bank and multiple classifiers to enhance the generalization performance of features. To ensure a balance between improved performance and computational efficiency, we retain only a subset of paths. The time consumption during testing in the Art Painting domain of the PACS on a single RTX 3090 is presented in Tab. \ref{tabr02}. It is evident that our method incurs only a slight increase in time compared to DeepAll~\cite{lv2023improving}, while achieving state-of-the-art performance.  Besides, as the classifier is solely a fully-connected layer, reflected in the results presented in Tab. \ref{tabr02}, our method experiences a slight increment in parameter count.

\begin{table}[htbp]
    \centering
    \caption{Parameter and Time results for various methods tested on  PACS.}
    \begin{tabular}{c|c|c|c}
        \toprule
        Methods & Parameters & Time per Image (ms/pic)  & Accuracy \\
        \midrule
       DeepAll \cite{lv2023improving} & 11.18M & 1.46   & 79.94  \\
       Mixstyle \cite{DBLP:conf/iclr/ZhouY0X21} & 11.18M  & 1.49  & 83.70 \\
        I$^2$-ADR \cite{DBLP:conf/eccv/MengLCYSWZSXP22} & 12.23M  & 1.44   & 85.55 \\
        \midrule
        Ours & 11.21M & 1.63   & 87.04  \\
        \bottomrule
    \end{tabular}%
    \label{tabr02}%
\end{table}

\textbf{Further evaluation of the different fusion scheme.}
We conduct experiments to investigate the influence of various prediction's strategies on generalization performance. We employed average and maximum fusion strategies for the independent domain paths and the main path, respectively. As shown in Tab. \ref{tabr03}, it is evident that our strategy achieves the highest average performance on four benchmark datasets. It shows that the averaging fusion strategy can improve generalization performance.  

\begin{table}[htbp]
  \centering
  \caption{Experimental results of different fusion schemes on prediction on PACS, Office-Home (OH), mini-DomainNet (mD), and  Digit\_DG (DD). ``M'' means the main path and ``I'' means the independent domain paths.}
    \begin{tabular}{l|cccc|c}
    \toprule
    \multicolumn{1}{c|}{Fusion Strategy} & ~PACS~  & OH & mD & DD & Avg.\\
    \midrule
    M & 86.30 & 65.45 & 66.80 & 84.48 & 75.76 \\
    Max(I) & 84.40 & 66.68 & 65.78 & 84.45 & 75.33\\
    Mean(I) & 84.02 & 66.21 & 67.40 & 84.38 & 75.50 \\
    Mean(I, M) & 86.21 & 67.70 & 67.88 & 85.33 & 76.78 \\
    Max(I, M) & 86.45 & 66.95 & 66.63 & 85.31 & 76.33 \\
    Max(Mean(I), M) & 86.58 & 66.45 & 67.28 & 85.34 & 76.41 \\
    Mean(Max(I), M) & \textbf{87.18} & \textbf{67.79} & 67.25 & 85.48 & 76.93 \\
    \midrule
    Ours Mean(Mean(I), M)  & 87.04 & \textbf{67.79} & \textbf{67.93} & \textbf{85.53} & \textbf{77.07}\\
    \bottomrule
    \end{tabular}%
  \label{tabr03}%
\end{table}%

\textbf{Visualization of the activation map.}
In this section, we present the activation maps of our method, as depicted in Fig.~\ref{fig03}. As observed, when compared to the baseline, our method exhibits a more focused activation on the key regions of each object. This highlights the effectiveness of our NormAUG method in improving the model's generalization in the target domain by capturing and highlighting the relevant features and regions.

\section{Conclusion}\label{s-conclusion}
In this paper, we introduce a novel data augmentation method based on the normalization perspective for domain generalization. Our method consists of two paths during the training stage: the main path, which serves as the baseline in the domain generalization task, and the auxiliary path, which incorporates our proposed augmentation method. Unlike traditional augmentation schemes, we adopt a random combination of domains for the normalization operation, thus enriching the diversity of the training data. Additionally, our method allows for the fusion of auxiliary results to enhance the model's performance in the test stage. Experimental results on various benchmark datasets demonstrate the effectiveness of our normalization-guided augmentation.

In our method, the normalization-based augmentation scheme does not introduce additional information, indicating that indirectly generating diverse style information in our method does not create a large domain gap between training samples, as shown in the theoretical analysis. However, the generated diverse data from our NormAUG is not the richest training set for each specific task. Therefore, in future work, we will explore introducing slight extra information to further enrich data's diversity while keeping the semantic information.

%
%

\ifCLASSOPTIONcaptionsoff
  \newpage
\fi

\bibliographystyle{IEEEtran}
\bibliography{sigproc}



 \vfill


\end{document}